\documentclass[runningheads]{llncs}

\usepackage{eccv}

\usepackage{eccvabbrv}

\usepackage{graphicx}
\usepackage{booktabs}

\usepackage[accsupp]{axessibility}  % Improves PDF readability for those with disabilities.

\usepackage{hyperref}
\usepackage{booktabs}
\usepackage{multicol}
\usepackage{colortbl}
\usepackage{color}
\usepackage{xcolor}
\usepackage{rotating}
\usepackage{multirow}
\usepackage{subcaption}
\usepackage{algorithm}
\usepackage{algorithmic}
\usepackage{marvosym}

\usepackage{orcidlink}

\begin{document}

% ---------------------------------------------------------------
% TODO REVIEW: Replace with your title
\title{SkelEM: Training-Signal Decoupling of Skeleton and Diffusion for 
Self-supervised Axial Super-Resolution in Volume Microscopy} 

% TODO REVIEW: If the paper title is too long for the running head, you can set
% an abbreviated paper title here. If not, comment out.
\titlerunning{SkelEM: Training-Signal Decoupling for Volume Microscopy ASR}

% TODO FINAL: Replace with your author list. 
% Include the authors' OCRID for the camera-ready version, if at all possible.
\author{Bohao Chen\inst{1,2}\orcidlink{0009-0002-4628-606X} \and
Yanchao Zhang\inst{2,3}\orcidlink{0009-0008-9313-5815} \and
Yanan Lv\inst{2,4} \and
Chenxun Deng\inst{2,4}\orcidlink{0009-0007-8423-9172} \and
Hua Han\inst{2,3,4}\orcidlink{0000-0003-4713-4631} \and
Xi Chen\inst{2,}\textsuperscript{\Letter}\orcidlink{0000-0002-6922-2838}
}

% TODO FINAL: Replace with an abbreviated list of authors.
\authorrunning{B.~Chen et al.}
% First names are abbreviated in the running head.
% If there are more than two authors, 'et al.' is used.

% TODO FINAL: Replace with your institution list.
\institute{School of Advanced Interdisciplinary Sciences, University of Chinese Academy of Sciences, China \and
State Key Laboratory of Brain Cognition and Brain-inspired Intelligence
Technology, Institute of Automation, Chinese Academy of Sciences, China \\
\email{\{chenbohao2024, zhangyanchao2021, lvyanan2018, \\ dengchenxun2025, hua.han, xi.chen\}@ia.ac.cn} \and
School of Future Technology, University of Chinese Academy of Sciences, China\\
\and
School of Artificial Intelligence, University of Chinese Academy of Sciences, China 
}

\maketitle

\begin{abstract}
Volume microscopy, including electron and light microscopy, suffers from severe anisotropic resolution due to physical axial sectioning. Existing self-supervised axial super-resolution (ASR) methods face a trilemma bounded by overly smoothed regression textures, structural hallucinations of pure diffusion models, and prohibitive inference latency. 
In this paper, we propose Skeleton-refinE Microscopy (SkelEM), a self-supervised framework that decouples ASR at the training-signal level: a frozen topological network and a diffusion refiner are optimized by disjoint objectives, separating low-frequency topology formulation from high-frequency detail enhancement.
Building on this deterministic skeleton, we exploit a unified cycle-consistent mechanism on input sparse slices to simultaneously extract a real-domain residual prior and bidirectionally align the diffusion refiner, washing away cross-plane artifacts without synthetic bias. By truncating the reverse diffusion process with this physical prior, SkelEM achieves high-fidelity detail restoration in merely $\le 5$ steps. To rigorously assess cross-instrument generalization, we further introduce BRAVE-ASR, a new benchmark of co-aligned anisotropic and isotropic volumes acquired on a Plasma-FIB instrument. Across public benchmarks, SkelEM achieves the most favorable balance across the fidelity-perception trade-off among self-supervised methods, with state-of-the-art downstream membrane segmentation performance and robust zero-shot generalization across distinct modalities.
\end{abstract}

\begin{figure}[t]
\centering
\includegraphics[width=1\linewidth]{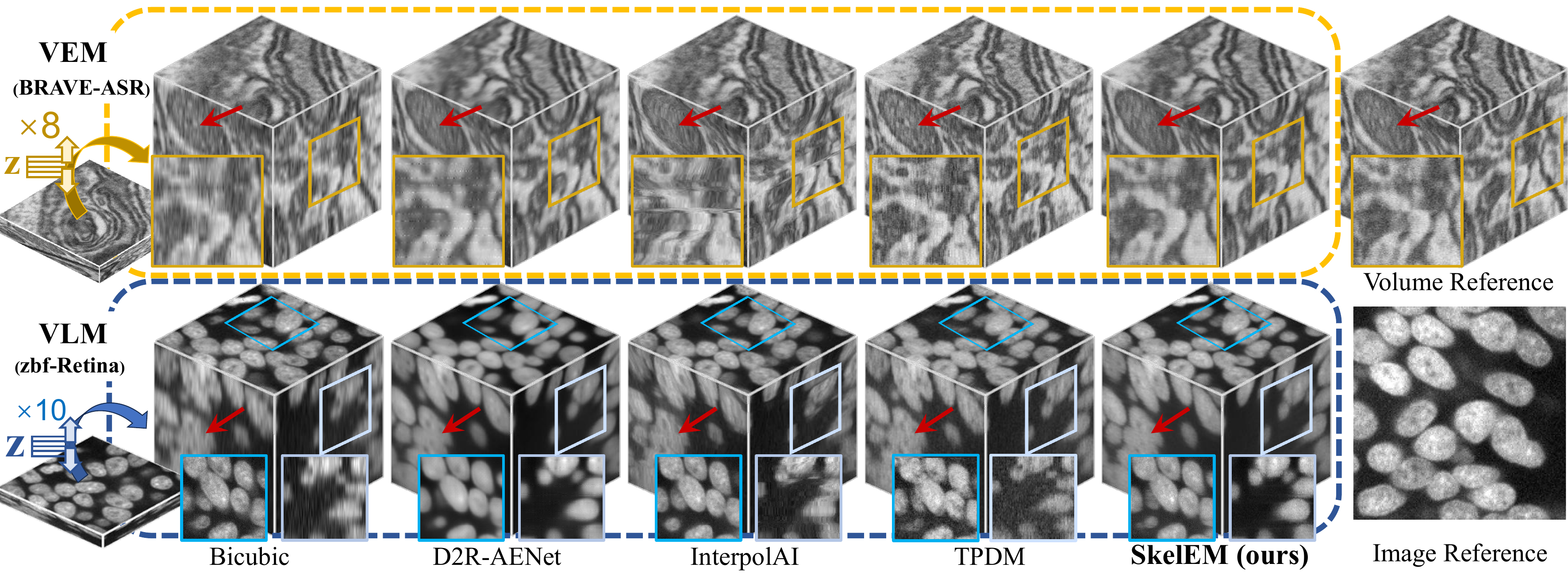}
\caption{SkelEM demonstrates robust axial super-resolution performance across different imaging modalities and instruments. 
\textbf{Top row (yellow box):} A challenging 8$\times$ axial super-resolution zero-shot instrument transfer task on our BRAVE-ASR dataset. \textbf{Bottom row (blue box):} A real-world (no high-resolution reference) 10$\times$ axial super-resolution task on a volume light microscopy dataset.}
\label{figure:first_vis}
\vspace{-1.5em}
\end{figure}

\section{Introduction}
\label{sec:intro}
Volume microscopy (VM) offers an unprecedented three-dimensional (3D) window 
into life's structure, from cellular organelles to complex tissue architectures
\cite{peddie2022volume, ma2022multiscale}. However, its effectiveness across 
both electron and light modalities is critically hampered by anisotropic 
resolution \cite{he2023isovem, ye2020axial}. Most high-throughput VM 
techniques, such as serial block-face scanning electron microscopy (SBF-SEM) 
or confocal light microscopy, are limited by physical sectioning or optical 
physics to an axial resolution far lower than their lateral resolution. For 
instance, SBF-SEM achieves a lateral resolution of 4-10 nm but is limited to 
25-40 nm axially \cite{borisovs2025new}, while standard confocal and 
two-photon microscopy are limited to 200-300 nm in lateral plane, versus 
500-800 nm axially \cite{elliott2020confocal}. While methods like focused ion 
beam scanning electron microscopy (FIB-SEM) can achieve isotropic nanoscale 
resolution \cite{xu2017enhanced}, their prohibitive cost and low throughput 
mean that the vast majority of large-volume datasets suffer from this axial 
deficiency \cite{kievits2022innovations}. The anisotropic resolution introduces 
structural and topological distortions \cite{schwartz2019removing} that can 
lead to erroneous circuit analysis in connectomics \cite{chen2024learning} and 
flawed morphological analysis in biology \cite{liu2022neuron}, 
directly impeding scientific discovery.

In recent years, computational approaches for axial super-resolution (ASR) 
have been developed to bridge this resolution gap, but face significant 
hurdles. Traditional interpolation produces blurry results, while supervised 
deep learning is impractical due to the extreme difficulty of acquiring 
isotropic 3D training volumes \cite{he2023isovem, lee2023reference}. 
Consequently, the field has shifted towards self-supervised methods. However, 
existing self-supervised paradigms face a fundamental fidelity-perception 
trade-off \cite{blau2018perception}. Regression-based 3D convolutional 
networks optimized via pixel-wise constraints inevitably suffer from 
regression-to-the-mean, producing overly smoothed volumes 
\cite{heinrich2017deep, chen2025self}, while generative methods such as 
diffusion models synthesize high-frequency details at the cost of structural 
hallucinations and prohibitive inference latency \cite{lee2023improving, 
lee2024reference, troidl2025niiv}. We attribute both failure modes to a 
shared root cause: the absence of a faithful structural skeleton that captures 
the highly nonlinear deformations of biological ultrastructures. Without such 
guidance, regression models over-smooth and generative models hallucinate. 
A domain-adapted structural skeleton would not only anchor reconstruction 
topology, but also enable truncated diffusion \cite{meng2021sdedit, 
yue2025arbitrary} to recover high-frequency details in very few steps, rendering 
the iterative denoising process both structurally reliable and computationally 
practical.

To this end, we propose Skeleton-refinE Microscopy (SkelEM), a self-supervised 
framework with two specialized stages decoupled at the training-signal level. SkelEM 
first establishes a deterministic structural skeleton via a topological network 
trained on domain-specific synthetic manifolds, deliberately discarding its 
detail refiner to enforce clean topology-texture separation. To recover missing 
biological textures without synthetic bias, a unified cycle-consistent mechanism 
on authentic sparse slices simultaneously extracts a real-domain residual prior 
and bidirectionally aligns a diffusion refiner to eliminate cross-plane artifacts. 
By initializing the reverse diffusion process from this physically grounded prior, 
SkelEM achieves high-fidelity reconstruction in merely $\leq 5$ steps while 
strictly preserving 3D structural integrity.

\begin{figure}[t]
\centering
\includegraphics[width=1\columnwidth]{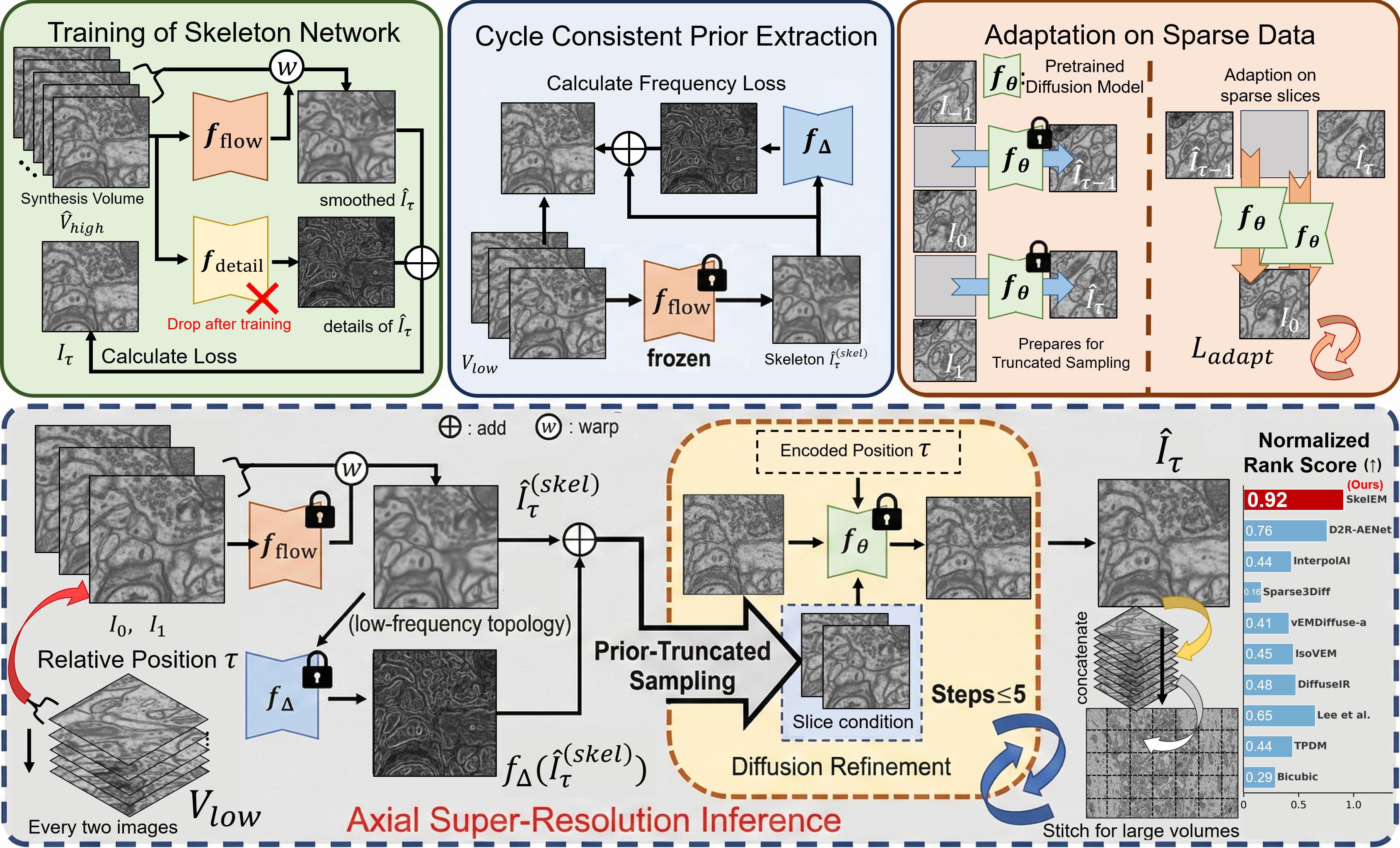}
\vspace{-1em}
\caption{Overview of the SkelEM framework. \textbf{Top: Three-stage training.} 
(\textit{Left}) $f_\text{flow}$ is trained on $\hat{V}_{high}$ to produce 
a smooth structural skeleton; $f_\text{detail}$ is deliberately discarded to 
enforce topology--texture separation. (\textit{Middle}) The frozen $f_\text{flow}$ 
cyclically reconstructs observed slices from $V_{low}$, and a residual estimator 
$f_\Delta$ is trained via frequency loss to enhance the high-frequency 
details. (\textit{Right}) The pretrained diffusion refiner $f_\theta$ is adapted on 
sparse real slices via bidirectional self-alignment loss $\mathcal{L}_{adapt}$ 
to eliminate synthetic bias and prepare for truncated sampling. 
\textbf{Bottom: Inference.} Given $I_0$, $I_1$ and target position $\tau$, 
the frozen $f_\text{flow}$ warps a structural skeleton $\hat{I}^{(skel)}_{\tau}$, 
and $f_\Delta$ predicts its high-frequency residual; their sum initializes the 
reverse diffusion at an intermediate timestep, from which $f_\theta$ completes 
refinement in $\leq 5$ steps. The bar chart on the right summarizes the 
normalized rank score $\bar{s}$ across 25 metrics on FIB-25, EPFL, BRAVE-ASR (zero-shot), 
and EPFL membrane segmentation; SkelEM attains $\bar{s}=0.92$, substantially 
ahead of all self-supervised baselines (see Suppl.~for details).}
\label{figure:main_figure}
\vspace{-1em}
\end{figure}

In summary, our contributions are: 
(1) SkelEM, a two-stage self-supervised ASR framework with training-signal decoupling—the topological skeleton network and the diffusion refiner are optimized by disjoint objectives with no shared gradients—where a domain-adapted structural skeleton serves as the key enabling condition for effective truncated diffusion in the microscopy domain.
(2) A unified cycle-consistent mechanism on authentic sparse slices that 
simultaneously eliminates synthetic bias and extracts a physically grounded 
residual prior, enabling high-fidelity diffusion refinement in $\leq 5$ steps. 
(3) Extensive validation on FIB-SEM benchmarks and the newly introduced BRAVE-ASR 
dataset demonstrates that SkelEM achieves the most favorable balance across 
the fidelity-perception trilemma among self-supervised methods (see normalized rank 
score in \cref{figure:main_figure}), delivering best perceptual quality in XZ/YZ 
views alongside competitive fidelity and SOTA downstream segmentation performance. The framework further demonstrates applicability beyond volume electron microscopy~(VEM), with qualitative validation on volume light microscopy~(VLM) confirming its modality-agnostic design.

\vspace{-1em}
\section{Related Work}
\label{sec:related}
\noindent \textbf{Axial Super-Resolution from Anisotropic Volumes.}
Enhancing axial resolution in anisotropic 3D volumes is a fundamental challenge
across VEM, VLM, and clinical imaging (e.g., CT/MRI). The extreme difficulty of
acquiring paired isotropic ground-truth volumes has driven the field toward
self-supervised methods~\cite{heinrich2017deep}, which leverage high-resolution
lateral (XY) planes as an internal supervisory signal to enhance the
low-resolution axial (Z) direction. Existing self-supervised ASR methods fall
into three paradigms, each with a characteristic failure mode. (1) 2D
super-resolution on orthogonal planes~\cite{weigert2017isotropic, deng2020isotropic, pan2023diffuseir, lee2023reference, troidl2025niiv, jiang2024super, park2022deep}, pioneered by Weigert \etal~\cite{weigert2017isotropic} and extended with unsupervised degradation learning~\cite{deng2020isotropic}, preserves lateral details but produces discontinuous cross-plane artifacts; (2) video frame
interpolation-inspired methods~\cite{jain2024video, lu2024diffusion,
joshi2025interpolai, wu2022computed} ensure 3D continuity but struggle with the
large nonlinear morphological changes of biological ultrastructures; (3) methods that refine 3D volumes by 2D diffusion models trained on XY planes ~\cite{lee2024reference, sun2025generalist,lee2023improving} achieve strong perceptual quality but incur prohibitive
computational costs~\cite{troidl2025niiv}. Navigating this trilemma between perceptual quality~\cite{blau2018perception}, 
3D structural consistency, and computational tractability remains an open 
challenge, motivating the training-signal decoupling strategy of our work.

\noindent \textbf{Video Frame Interpolation.}
Video frame interpolation (VFI) generates intermediate frames between existing 
input frames and serves as a strong methodological analogue for ASR. A dominant 
paradigm, exemplified by Super-SloMo~\cite{jiang2018super} and 
RIFE~\cite{huang2022real}, explicitly estimates optical flow to warp and 
synthesize intermediate frames and continues to be refined for challenging 
non-linear motions~\cite{liu2020enhanced}. Complementing these flow-based 
approaches, recent flow-free methods leverage state-space models such as 
Mamba~\cite{jeong2025lc} or diffusion-based models~\cite{jain2024video, 
hai2025hierarchical}. While we adopt the flow estimation backbone of 
RIFE~\cite{huang2022real} as our topological skeleton network, a critical 
distinction from standard VFI pipelines is that we deliberately discard 
its detail refiner after pre-training. This design choice prevents the leakage of 
synthetic hallucinations inherent in pseudo-HR volumes into the texture generation 
stage, enforcing a strict separation between low-frequency topology and 
high-frequency detail recovery. Furthermore, as we demonstrate in 
\cref{sec:ablation}, VFI models pre-trained on natural videos fail to capture 
the complex nonlinear deformations of biological ultrastructures, necessitating 
domain-specific re-training on synthetic microscopy manifolds.

\noindent \textbf{Diffusion Models for Image Restoration.}
Diffusion probabilistic models (DPMs) have achieved state-of-the-art results in 
high-fidelity image restoration~\cite{ho2020denoising, dhariwal2021diffusion}, 
but their iterative denoising process renders them impractical for large-scale 
biological volumes~\cite{troidl2025niiv}. While truncated diffusion 
strategies~\cite{song2020denoising, yue2025arbitrary} can accelerate inference 
by initializing from an informative prior, their effectiveness critically depends 
on the quality of that initialization. In the absence of a structurally reliable 
prior, such truncation merely shifts rather than resolves the hallucination 
problem. This motivates our two-stage design, where a domain-specific 
deterministic skeleton serves as the structural anchor for truncated diffusion, 
enabling high-fidelity synthesis in $\leq$5 steps.

\vspace{-1em}

\section{Proposed Method}
\label{sec:method}

Let $V_{low} \in \mathbb{R}^{d \times h \times w}$ denote the acquired anisotropic 
volume with high-resolution lateral dimensions and $d$ sparse slices along the 
axial (Z) direction. Axial super-resolution (ASR) aims to reconstruct an axially 
high-resolution volume $V_{high} \in \mathbb{R}^{\hat{d} \times h \times w}$ by 
synthesizing intermediate slices, where the upsampling factor is $r \ge 2$ and 
$\hat{d} = (d-1) \times r + 1$. For any two adjacent slices $I_0$ and $I_1$ in 
$V_{low}$, the objective is to generate the missing intermediate slice at a 
relative position $\tau \in (0, 1)$. As established by the fidelity-perception trade-off~\cite{blau2018perception}, directly learning an end-to-end mapping $\mathcal{F}: \{I_0, I_1, \tau\} 
\rightarrow \hat{I}_\tau$ is highly ill-posed, inevitably collapsing toward 
either regression smoothing or generative hallucinations.

SkelEM resolves this through training-signal decoupling: the topological network and the diffusion refiner are optimized by disjoint objectives with no shared gradients (illustrated in \cref{figure:main_figure}). Unlike cascaded pipelines that share supervision across the topology and texture stages~\cite{lee2024reference, pan2023diffuseir}, this separation prevents synthetic-texture leakage into the structural skeleton while enabling each stage to specialize. First, a topological network trained on synthetic manifolds establishes a deterministic skeleton as a stable low-frequency anchor (\cref{sec:skeleton}); a diffusion refiner then recovers high-frequency biological textures, trained on synthetic manifolds and self-aligned on sparse real slices (\cref{sec:alignment}). A lightweight residual estimator bridges the two via prior-truncated sampling (\cref{sec:inference}).

\vspace{-1em}
\subsection{Topological Skeleton via Synthetic Manifold Pre-training}
\label{sec:skeleton}

Biological ultrastructures exhibit highly non-linear topological deformations 
across the large axial gaps of $V_{low}$. Although flow-based VFI networks 
provide a natural structural prior for slice synthesis, models pre-trained on 
natural videos struggle to capture the complex nonlinear dynamics of biological 
tissues, as we demonstrate in \cref{sec:ablation}. This domain gap motivates 
training a flow network on synthetic microscopy manifolds rather than 
directly using natural video priors~\cite{jain2024video, joshi2025interpolai}.

To this end, we leverage a prior distillation method~\cite{chen2025self} to 
generate a dense pseudo-HR volume $\hat{V}_{high}$ as training data, and adopt 
the intermediate flow network (IFNet) from RIFE~\cite{huang2022real} as our 
topological network $f_{flow}$. Given adjacent authentic slices $I_0$ and $I_1$ 
at a relative position $\tau \in (0, 1)$, $f_{flow}$ predicts bi-directional 
flows $F_{\tau \rightarrow 0}, F_{\tau \rightarrow 1}$ and a blending mask 
$M_\tau \in [0, 1]$. The structural skeleton $\hat{I}^{(skel)}_{\tau}$ is synthesized via backward warping $\mathcal{W}$:
\begin{equation}
\hat{I}^{(skel)}_{\tau} = M_\tau \odot \mathcal{W}(I_0, F_{\tau \rightarrow 0}) + 
(1 - M_\tau) \odot \mathcal{W}(I_1, F_{\tau \rightarrow 1})
\end{equation}

As discussed in \cref{sec:related}, standard VFI pipelines append a detail 
refiner to inject high-frequency textures. Since $\hat{V}_{high}$ inherently 
contains cross-plane artifacts, however, retaining this refiner would inevitably 
leak synthetic hallucinations into the skeleton. We therefore deliberately discard 
the detail refiner after training on $\hat{V}_{high}$ and strictly freeze $f_{flow}$, restricting 
it to function exclusively as a deterministic low-frequency topological anchor. 
This enforces the central separation principle of our framework: $f_{flow}$'s gradients never reach the refinement stage, isolating spatial warping from the subsequent real-domain texture recovery described in \cref{sec:alignment}. The quality and stability of $\hat{V}_{high}$ as a training 
signal is empirically corroborated by D2R-AENet~\cite{chen2025self}, which trains 
exclusively on the same synthetic volume and consistently achieves the highest 
fidelity scores among all self-supervised baselines (see \cref{tab:quantitative}).

\subsection{Diffusion Detail Refiner and Sparse Self-Alignment}
\label{sec:alignment}

While the topological skeleton $\hat{I}^{(skel)}_{\tau}$ establishes a robust 
low-frequency foundation, it inherently lacks the fine-grained biological 
textures lost during physical sectioning. To restore these missing high-frequency 
details, we employ a dedicated diffusion-based refinement network $f_{\theta}$ 
that learns to enhance the intermediate high-frequency distribution conditioned 
on boundary slices $I_0$ and $I_1$.

\noindent\textbf{Base Training on Synthetic Manifolds.}
We first establish a base texture generation capability by training $f_{\theta}$ 
on the pseudo-HR volume $\hat{V}_{high}$ from \cref{sec:skeleton}. For an 
intermediate target slice $\hat{I}_\tau$ sampled from the synthetic volume, the 
model is optimized via the standard diffusion objective:
\begin{equation}
\mathcal{L}_{base} = \mathbb{E}_{t, \hat{I}_\tau, \epsilon} \left[ 
\| \epsilon - f_{\theta}(x_{t}, t, I_0, I_1, \tau) \|_2^2 \right],
\end{equation}
where $t$ is the diffusion timestep and $x_{t}$ represents the noisy state of 
$\hat{I}_\tau$. This phase equips the model with a strong generative prior 
capturing the macroscopic biological score function without relying on massive 
external pre-trained models. However, since $\hat{V}_{high}$ inherently carries 
cross-plane artifacts, the refiner trained solely on synthetic data inevitably 
inherits these biases, motivating the subsequent sparse self-alignment stage.

\noindent\textbf{Sparse Self-Alignment on Real Slices.}
To eliminate the synthetic bias accumulated during base training, we fine-tune 
$f_{\theta}$ directly on the authentic sparse slices of $V_{low}$. Given a 
sliding window of three slices $\{I_{-1}, I_0, I_1\}$, we first synthesize a 
pair of intermediate anchors using the current refiner: $\hat{I}_{\tau-1}$ 
(generated between $I_{-1}$ and $I_0$) and $\hat{I}_{\tau}$ (generated between 
$I_0$ and $I_1$). Unlike Sparse3Diff~\cite{lee2025sparse3diff}, which relies on 
external pre-trained models and applies only a unidirectional constraint, we 
optimize $f_{\theta}$ via a unified \textit{bidirectional} self-alignment 
objective. By alternating the conditioning sequence and its complementary 
geometric index, the model is required to reconstruct the central observed slice 
$I_0$ consistently regardless of synthesis direction, enforcing topological 
symmetry and suppressing directional bias:
\begin{equation}
\mathcal{L}_{adapt} = \sum_{\phi \in \Phi} \mathbb{E}_{\epsilon, I_0, t} \left[ 
\| \epsilon - f_{\theta}(x_{t}, t, \phi) \|_2^2 \right],
\end{equation}
where $\Phi = \{(\hat{I}_{\tau-1}, \hat{I}_\tau, 1-\tau), 
(\hat{I}_\tau, \hat{I}_{\tau-1}, \tau)\}$, and the synthesized anchors 
$\hat{I}_{\tau-1}$, $\hat{I}_{\tau}$ are treated as fixed conditioning inputs 
without gradient backward during the optimization of $\mathcal{L}_{adapt}$. 
Intuitively, a direction-agnostic refiner should reconstruct $I_0$ equally well from either temporal direction.

\subsection{Fast Inference via Prior-Truncated Sampling}
\label{sec:inference}

Standard diffusion requires hundreds of steps from Gaussian noise, rendering it
computationally prohibitive for gigapixel biological volumes. To achieve highly
accelerated inference without compromising fidelity, we propose a prior-truncated
sampling strategy enabled by a real-domain high-frequency residual prior.

\noindent\textbf{Cycle-Consistent Prior Extraction.}
Building upon the topological symmetry established in \cref{sec:alignment}, we
formulate a deterministic cyclic reconstruction process utilizing the frozen
$f_{flow}$. Operating within the slice sequence $\{I_{-1}, I_0, I_1\}$ from
$V_{low}$, we warp two virtual anchors at symmetric topological positions:
$I_{-\tau}$ and $I_{1-\tau}$. The central slice is cyclically reconstructed via
\begin{equation}
    \tilde{I}_0 = \mathcal{W}_{cycle}(I_{-\tau}, I_{1-\tau}, f_{flow}, \tau).
\end{equation}
Their difference $\Delta_{gt} = I_0 - \tilde{I}_0$ explicitly isolates the
high-frequency details lost during nonlinear warping. To generalize this
extraction to unobserved slices, a lightweight estimator $f_{\Delta}$ is trained
to predict this residual directly from the structural skeleton. To emphasize
high-frequency fidelity, the estimator is supervised by the frequency L1 loss:
\begin{equation}
    \mathcal{L}_{freq} = \| \mathcal{F}_{fft}(f_{\Delta}(\hat{I}^{(skel)}_{\tau})) -
    \mathcal{F}_{fft}(\Delta_{gt}) \|_1,
\end{equation}
where $\mathcal{F}_{fft}(\cdot)$ denotes the fast Fourier transform.

\noindent\textbf{Initialization via Prior Truncation.}
As established in \cref{sec:related}, truncated diffusion strategies are
effective only when initialized from a structurally reliable prior. The
cycle-consistent residual $\Delta_\tau = f_{\Delta}(\hat{I}^{(skel)}_{\tau})$ extracted
above provides precisely such an anchor. For an unobserved target slice at
relative position $\tau$, we aggregate the deterministic skeleton
$\hat{I}^{(skel)}_{\tau}$ and $\Delta_\tau$ into a prior state
$\hat{I}^{(prior)}_{\tau} = \hat{I}^{(skel)}_{\tau} + \Delta_\tau$, where the predicted
residual belongs to pixel space rather than noise space. Rather than initiating
the reverse process from a random state $x_T \sim \mathcal{N}(0, I)$, we
truncate the generative trajectory to an intermediate timestep $t_S$ ($S \ll T$),
formulating the initial latent state as:
\begin{equation}
\label{eq:init_state_inversion}
x_{t_S} = \sqrt{\bar{\alpha}_{t_S}} \left( \hat{I}^{(skel)}_{\tau} + \Delta_\tau
\right) + \sqrt{1 - \bar{\alpha}_{t_S}} \epsilon,
\end{equation}
where $\bar{\alpha}_{t_S}$ dictates the noise schedule and $\epsilon \sim
\mathcal{N}(0, I)$ is standard Gaussian noise. This forward-simulated
initialization anchors the reverse trajectory to a structurally correct manifold
already enriched with authentic high-frequency cues, substantially reducing the
generative burden. We acknowledge that this initialization is a heuristic
approximation rather than an exact inversion, and validate its effectiveness
through ablation studies in \cref{sec:ablation}.

\noindent\textbf{Rapid Iterative Refinement.}
Starting from the prior-injected state $x_{t_S}$, we perform $S$ steps of
reverse sampling ($S \le 5$) using the bidirectionally aligned refiner
$f_{\theta}$ from \cref{sec:alignment}. Defining a sequence of diffusion
timesteps $\{t_S, t_{S-1}, \dots, t_0=0\}$, for each step $n$ traversing from
$S$ down to $1$, the denoising model predicts the clean state $\hat{x}_0$
conditioned on the authentic boundary slices $I_0$, $I_1$ and the spatial
index $\tau$:
\begin{equation}
\label{eq:reverse_step_pred_x0}
\hat{x}_0 = \frac{1}{\sqrt{\bar{\alpha}_{t_n}}} \left( x_{t_n} -
\sqrt{1 - \bar{\alpha}_{t_n}} f_{\theta}(x_{t_n}, t_n, I_0, I_1, \tau)
\right).
\end{equation}
The subsequent latent state $x_{t_{n-1}}$ is then computed following the
standard DDPM reverse formulation:
\begin{equation}
\label{eq:reverse_step_ddpm}
x_{t_{n-1}} = \frac{\sqrt{\alpha_{t_n}}(1 -
\bar{\alpha}_{t_{n-1}})}{1 - \bar{\alpha}_{t_n}}x_{t_n} +
\frac{\sqrt{\bar{\alpha}_{t_{n-1}}}(1 -
\alpha_{t_n})}{1 - \bar{\alpha}_{t_n}}\hat{x}_0 + \sigma_{t_n}z.
\end{equation}

\vspace{-1em}
\section{Experiments}
\label{sec:experiments}

\subsection{Datasets and Implementation Details}
\label{sec:datasets_implementation}

\subsubsection{Datasets and Degradation Process}
\label{sec:datasets}
To comprehensively evaluate our framework, we utilize three public volume microscopy datasets (FIB-25 \cite{takemura2015synaptic}, EPFL \cite{epfldataset}, and a Zebrafish retina VLM dataset \cite{weigert2018content}) alongside our newly proposed BRAVE-ASR dataset. While FIB-25 and EPFL (acquired via FIB-SEM) serve as our primary training and in-domain testing grounds, we introduce Benchmarking Anisotropic Volume-microscopy Evaluation for Axial Super-Resolution (BRAVE-ASR, acquired via Plasma-FIB) specifically to establish a standardized benchmark for zero-shot instrument transfer (further detailed in \cref{sec:zero_shot_comparison}). To simulate the anisotropic degradation, we adopt the z-axis downsampling strategy from vEMDiffuse \cite{lu2024diffusion} with a factor of $r=8$. This strategy retains the intrinsic physical noise of the original sections, faithfully simulating the anisotropic acquisition process where axial slices are physically sectioned at sparse intervals. All datasets are split into train/validation/test sets following a 70\%/15\%/15\% ratio. Comprehensive details regarding specimens, spatial resolutions, and the quantitative validation of the above downsampling strategy via BRAVE-ASR dataset are provided in the Supplementary Materials.

\subsubsection{Training Details}
\label{sec:training_details}
All networks were trained and evaluated on patches of $256\times256$
pixels by using the ADAM optimizer ($\beta_1=0.9, \beta_2=0.99$). The skeleton network was trained on $\hat{V}_{high}$ for 300 epochs by L1 loss, requiring approximately 12 hours on a single GPU. The diffusion refiner underwent two-phase training: (1) pre-training on $\hat{V}_{high}$ for 880 epochs, followed by (2) self-alignment fine-tuning ($\mathcal{L}_{\text{adapt}}$) on the sparse $V_{low}$ for 20 epochs. We use a ResNet-9 as the detail estimator and train it for 5 epochs. All components took approximately 72 hours using two GPUs for full training. The experiments were conducted using PyTorch 2.4.1 on a Linux server equipped with three Nvidia 4090 GPUs.

\subsubsection{Comparison with State-of-the-art Methods}
\label{sec:comparison_sota}
We benchmark our method against bicubic interpolation and a comprehensive suite of state-of-the-art self-supervised ASR methods, with supervised methods (SRUNet~\cite{heinrich2017deep}, vEMDiffuse-i~\cite{lu2024diffusion}) included for reference. Performance is evaluated using 3D-PSNR, SSIM, and LPIPS~\cite{zhang2018unreasonable}. All methods were trained following their official implementations, and evaluated on $256^2$ patches following standard practice~\cite{lee2023improving, lu2024diffusion, lee2024reference, pan2023diffuseir}. 

\subsection{Quantitative Performance Comparison}
\label{sec:quant_comparison}

\begin{table}[t]
\centering
\footnotesize
\caption{Quantitative evaluation on FIB25 and EPFL datasets.
Best/second-best results in self-supervised methods are marked in \textbf{bold}/\underline{underlined}, respectively.
Supervised (Sup.) methods are listed for reference only and not included in the ranking.}
\begin{tabular}{c|c|c|c|ccc|ccc}
\toprule
\multirow{2}{*}{\rotatebox{90}{Dataset}} & \multirow{2}{*}{\rotatebox{90}{Type}} & \multirow{2}{*}{Methods}
& \multirow{2}{*}{3D-PSNR($\uparrow$)}
& \multicolumn{3}{c|}{SSIM($\uparrow$)}
& \multicolumn{3}{c}{LPIPS($\downarrow$)} \\
\cmidrule{5-10}
& & & & XY & XZ & YZ & XY & XZ & YZ \\
\midrule
% FIB25数据集（修正行列对齐）
\multirow{12}{*}{\rotatebox{90}{FIB25}}
& \multirow{2}{*}{\rotatebox{90}{Sup.}}
& SRUNet~\cite{heinrich2017deep}          & 26.71 & 0.7168 & 0.7119 & 0.7061 & 0.3718 & 0.4371 & 0.4525 \\
& & vEMDiffuse-i~\cite{lu2024diffusion}    & 25.68 & 0.6669 & 0.6287 & 0.6210 & 0.2653 & 0.3994 & 0.4380 \\
\cline{2-10}
& \multirow{10}{*}{\rotatebox{90}{Self-sup.}}
& Bicubic                                & 25.33 & \underline{0.6528} & 0.6339 & 0.6254 & \textbf{0.2578} & 0.5881 & 0.5971 \\
& & TPDM~\cite{lee2023improving}           & 23.38 & 0.5129 & 0.5186 & 0.5109 & 0.3613 & 0.4448 & 0.4203 \\
& & Lee~et~al.~\cite{lee2024reference}     & 26.18 & 0.6607 & 0.6529 & 0.6456 & 0.3332 & \underline{0.3862} & \underline{0.3760} \\
& & DiffuseIR~\cite{pan2023diffuseir}      & 25.13 & 0.5820 & 0.5724 & 0.5663 & 0.4131 & 0.4394 & 0.4382 \\
& & IsoVEM~\cite{he2023isovem}             & 24.57 & 0.6350 & 0.6329 & 0.6226 & \underline{0.2654} & 0.5296 & 0.5578 \\
& & vEMDiffuse-a~\cite{lu2024diffusion}    & 23.51 & 0.5699 & 0.5244 & 0.5164 & 0.3455 & 0.4260 & 0.4466 \\
& & Sparse3Diff~\cite{lee2025sparse3diff}  & 22.83 & 0.5568 & 0.4131 & 0.4042 & 0.3265 & 0.5664 & 0.6056 \\
& & InterpolAI~\cite{joshi2025interpolai}  & 24.27 & 0.6140 & 0.5676 & 0.5613 & 0.2943 & 0.5535 & 0.5819 \\
& & D2R-AENet~\cite{chen2025self}          & \textbf{27.62} & \textbf{0.7373} & \textbf{0.7271} & \textbf{0.7206} & 0.3556 & 0.4136 & 0.4246 \\
\cline{3-10}
& & SkelEM (ours)                         & \underline{26.24} & \underline{0.6765} & \underline{0.6547} & \underline{0.6478} & 0.2721 & \textbf{0.3658} & \textbf{0.3687} \\
\midrule
% EPFL数据集（修正行列对齐+更新引用）
\multirow{12}{*}{\rotatebox{90}{EPFL}}
& \multirow{2}{*}{\rotatebox{90}{Sup.}}
& SRUNet~\cite{heinrich2017deep}          & 25.61 & 0.6476 & 0.6863 & 0.6813 & 0.5272 & 0.4678 & 0.4717 \\
& & vEMDiffuse-i~\cite{lu2024diffusion}    & 24.45 & 0.5514 & 0.5833 & 0.5779 & 0.2652 & 0.3034 & 0.3094 \\
\cline{2-10}
& \multirow{10}{*}{\rotatebox{90}{Self-sup.}}
& Bicubic                                & 23.09 & 0.4938 & 0.5191 & 0.5100 & \underline{0.3012} & 0.6794 & 0.6674 \\
& & TPDM~\cite{lee2023improving}           & 22.82 & 0.4976 & 0.5279 & 0.5210 & \textbf{0.2859} & 0.4070 & 0.4029 \\
& & Lee~et~al.~\cite{lee2024reference}     & 24.99 & 0.5593 & 0.6044 & 0.5974 & 0.3428 & \underline{0.3327} & \underline{0.3431} \\
& & DiffuseIR~\cite{pan2023diffuseir}      & 24.48 & 0.5059 & 0.5625 & 0.5560 & 0.3982 & 0.3649 & 0.3788 \\
& & IsoVEM~\cite{he2023isovem}             & 22.98 & 0.5085 & 0.5542 & 0.5436 & \underline{0.3012} & 0.6434 & 0.6528 \\
& & vEMDiffuse-a~\cite{lu2024diffusion}    & 22.98 & 0.4879 & 0.5092 & 0.5007 & 0.3808 & 0.4269 & 0.4060 \\
& & Sparse3Diff~\cite{lee2025sparse3diff}  & 22.05 & 0.4723 & 0.4516 & 0.4438 & 0.3752 & 0.6022 & 0.6063 \\
& & InterpolAI~\cite{joshi2025interpolai}  & 23.67 & 0.5116 & 0.5426 & 0.5376 & 0.3491 & 0.5727 & 0.5861 \\
& & D2R-AENet~\cite{chen2025self}          & \textbf{26.35} & \textbf{0.6387} & \textbf{0.6793} & \textbf{0.6734} & 0.4764 & 0.4788 & 0.4795 \\
\cline{3-10}
& & SkelEM (ours)                         & \underline{25.56} & \underline{0.6192} & \underline{0.6532} & \underline{0.6464} & 0.3049 & \textbf{0.3213} & \textbf{0.3298} \\
\bottomrule
\end{tabular}
\label{tab:quantitative}
\vspace{-1.5em}
\end{table}

We applied SkelEM and comparative methods to reconstruct volumes from the FIB-25 and EPFL test datasets, calculating image similarity metrics in all three views, as shown in \cref{tab:quantitative}. The results directly reflect the trilemma outlined in \cref{sec:intro}: self-supervised methods are broadly trapped by a fidelity-versus-perception trade-off, with no single baseline resolving all three axes simultaneously. Regression-based networks like D2R-AENet achieve the highest fidelity scores at the cost of severe perceptual degradation, confirming their over-smoothing tendency. Other diffusion-based methods either compromise 3D structural consistency or overall fidelity under self-supervised constraints. In contrast, SkelEM resolves this trilemma through training-signal decoupling between its two specialized stages: it consistently achieves the best perceptual quality in the XZ/YZ views, secures highly competitive fidelity scores across both datasets, and as demonstrated in \cref{tab:abl_steps}, requires merely 3 inference steps—over an order of magnitude faster than existing diffusion-based ASR methods. This unique balance across all three axes validates that dedicated separation of structural continuity and high-frequency detail synthesis is essential for practical, high-fidelity isotropic reconstruction.

\subsubsection{Zero-Shot Instrument Transfer Comparison}
\label{sec:zero_shot_comparison}

\begin{table}[t]
\centering
\footnotesize
\caption{Quantitative evaluation of zero-shot domain transfer on the BRAVE-ASR dataset. All models were trained on the EPFL dataset and evaluated directly on BRAVE-ASR without fine-tuning. Best/second-best results in self-supervised (Self-sup.) methods are marked in \textbf{bold}/\underline{underlined}, respectively. Supervised (Sup.) methods are listed for reference and not included in the ranking.}
\begin{tabular}{c|c|c|c|ccc|ccc}
\toprule
\multirow{2}{*}{\rotatebox{90}{Dataset}} & \multirow{2}{*}{\rotatebox{90}{Type}} & \multirow{2}{*}{Methods}
& \multirow{2}{*}{3D-PSNR($\uparrow$)}
& \multicolumn{3}{c|}{SSIM($\uparrow$)}
& \multicolumn{3}{c}{LPIPS($\downarrow$)} \\
\cmidrule{5-10}
& & & & XY & XZ & YZ & XY & XZ & YZ \\
\midrule
\multirow{11}{*}{\rotatebox{90}{BRAVE-ASR}}
& \multirow{2}{*}{\rotatebox{90}{Sup.}}
& SRUNet~\cite{heinrich2017deep}          & 23.67 & 0.5979 & 0.6274 & 0.6325 & 0.5210 & 0.4923 & 0.4810 \\
& & vEMDiffuse-i~\cite{lu2024diffusion}    & 21.99 & 0.4777 & 0.4902 & 0.4965 & 0.3920 & 0.4257 & 0.3943 \\
\cline{2-10}
& \multirow{9}{*}{\rotatebox{90}{Self-sup.}}
& Bicubic                               & 20.44 & 0.4352 & 0.4069 & 0.4147 & \underline{0.3726} & 0.7236 & 0.7166 \\
& & TPDM~\cite{lee2023improving}           & 21.02 & 0.4489 & 0.4598 & 0.4668 & 0.3769 & 0.4570 & 0.4203 \\
& & Lee~et~al.~\cite{lee2024reference}     & 21.12 & 0.3998 & 0.4291 & 0.4354 & 0.3926 & \underline{0.3634} & \underline{0.3492} \\
& & DiffuseIR~\cite{pan2023diffuseir}       & 22.40 & 0.4661 & 0.4584 & 0.4673 & 0.4404 & 0.4177 & 0.4076 \\
& & IsoVEM~\cite{he2023isovem}             & 23.23 & 0.5318 & 0.5571 & 0.5612 & 0.4409 & 0.6009 & 0.5894 \\
& & vEMDiffuse-a~\cite{lu2024diffusion}    & 21.73 & 0.4729 & 0.4825 & 0.4893 & 0.4082 & 0.4346 & 0.3853 \\
& & Sparse3Diff~\cite{lee2025sparse3diff}  & 21.35 & 0.4677 & 0.4451 & 0.4498 & 0.4027 & 0.5883 & 0.5559 \\
& & InterpolAI~\cite{joshi2025interpolai}  & 21.58 & 0.4472 & 0.4658 & 0.4728 & \textbf{0.3695} & 0.6475 & 0.6337 \\
& & D2R-AENet~\cite{chen2025self}          & \textbf{24.15} & \textbf{0.5814} & \textbf{0.6116} & \textbf{0.6124} & 0.4802 & 0.5351 & 0.5109 \\
\cline{3-10}
& & SkelEM (ours)                         & \underline{23.48} & \underline{0.5665} & \underline{0.5892} & \underline{0.5935} & 0.3747 & \textbf{0.3203} & \textbf{0.3093} \\
\bottomrule
\end{tabular}
\label{tab:BRAVE_table}
\vspace{-1.5em}
\end{table}

\begin{figure}[t!]
\centering
\footnotesize
\includegraphics[width=1\columnwidth]{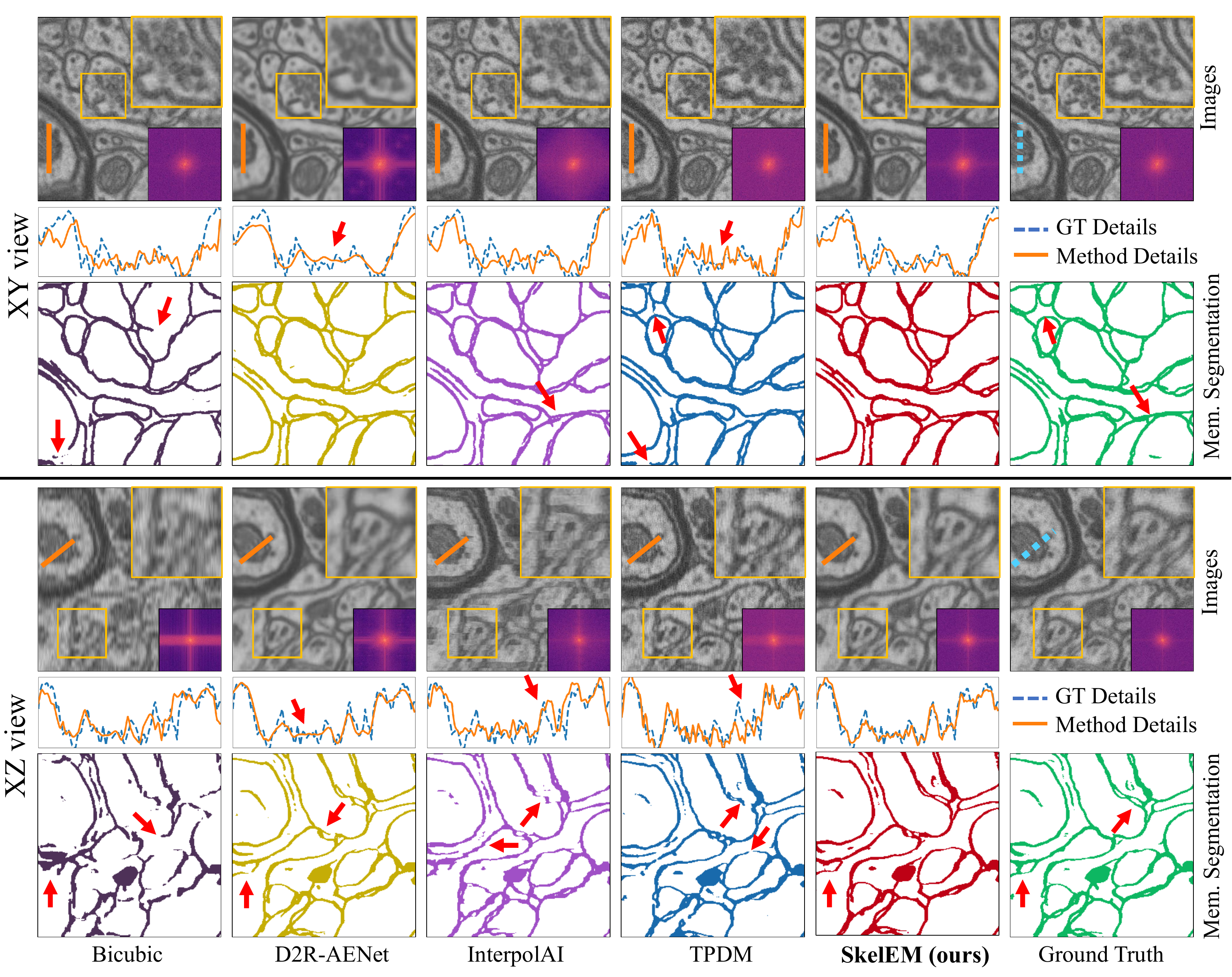} 
\caption{Qualitative comparison of SkelEM with state-of-the-art self-supervised ASR methods on the EPFL dataset ($r=8$) across two orthogonal views (XY and XZ). 
For each view, we show the reconstructed slices with zoomed-in detail insets, 
frequency spectra, and intensity profiles (orange) compared against the ground-truth 
(blue dashed), alongside the corresponding 3D membrane segmentation results. 
Red arrows highlight topological errors such as broken membranes and spurious connections. 
SkelEM (ours) best preserves high-frequency structural details, yielding membrane 
segmentations and details most consistent with the ground-truth among all others.}
\label{figure:seg_fig}
\vspace{-1.5em}
\end{figure}

\begin{table}[ht]
\footnotesize
\centering
\caption{Membrane segmentation accuracy on the EPFL test set ($r=8$). Best/second-best results in self-supervised methods are marked in \textbf{bold}/\underline{underlined}, respectively. Supervised (Sup.) methods are listed for reference and not included in the ranking.}
\begin{tabular}{c|c|cccc}
\toprule
\multirow{2}{*}{\rotatebox{90}{Type}} & \multirow{2}{*}{Method} & \multicolumn{4}{c}{Membrane Segmentation} \\
\cmidrule{3-6}
& & F1 Score($\uparrow$) & IoU ($\uparrow$) & ARE ($\downarrow$) & VoI ($\downarrow$) \\
\midrule

\multirow{2}{*}{\rotatebox{90}{Sup.}}
& SRUNet~\cite{heinrich2017deep}          & 0.8233 & 0.6996 & 0.2707 & 0.5667 \\
& vEMDiffuse-i~\cite{lu2024diffusion}    & 0.8610 & 0.7560 & 0.2179 & 0.4896 \\

\midrule

\multirow{11}{*}{\rotatebox{90}{Self-sup.}}
& Bicubic                               & 0.6969 & 0.5348 & 0.4380 & 0.7839 \\
& TPDM~\cite{lee2023improving}           & 0.8391 & 0.7228 & 0.2511 & 0.5464 \\
& Lee~et~al.~\cite{lee2024reference}     & 0.8319 & 0.7121 & 0.2568 & 0.5389 \\
& DiffuseIR~\cite{pan2023diffuseir}      & 0.7885 & 0.6508 & 0.3139 & 0.6078 \\
& IsoVEM~\cite{he2023isovem}             & 0.7926 & 0.6564 & 0.3172 & 0.6463 \\
& vEMDiffuse-a~\cite{lu2024diffusion}    & 0.8015 & 0.6687 & 0.3035 & 0.6229 \\
& Sparse3Diff~\cite{lee2025sparse3diff}  & 0.6829 & 0.5185 & 0.4645 & 0.8434 \\
& InterpolAI~\cite{joshi2025interpolai}  & 0.8058 & 0.6747 & 0.2962 & 0.6080 \\
& D2R-AENet~\cite{chen2025self}          & \underline{0.8564} & \underline{0.7489} & \underline{0.2240} & \underline{0.4974} \\
\cline{2-6}
& SkelEM (ours)                         & \textbf{0.8643} & \textbf{0.7611} & \textbf{0.2129} & \textbf{0.4811} \\
\bottomrule
\end{tabular}
\label{tab:downstream_segmentation}
\vspace{-1em}
\end{table}

Generalization across disparate microscopy hardware remains a fundamental bottleneck in microscopy. To address this, a key contribution of our work is establishing a standardized zero-shot instrument transfer benchmark using the proposed BRAVE-ASR dataset. Specifically, we evaluate models trained on EPFL (Zeiss NVision40 FIB-SEM) directly on the unseen BRAVE-ASR testbed (Thermo Scientific Helios Hydra Plasma FIB-SEM). As shown in \cref{figure:first_vis} and \cref{tab:BRAVE_table}, existing SOTA methods struggle significantly with this domain shift. VFI-based methods such as InterpolAI~\cite{joshi2025interpolai} 
suffer from a natural-to-biological domain gap; while reusing 
input textures preserves surface-level perception, it introduces 
severe structural distortions and degrades axial fidelity. Pure diffusion models achieve higher perception but fail on fidelity metrics. Conversely, 3D convolutional networks like D2R-AENet~\cite{chen2025self} preserve high fidelity but produce overly smooth results with degraded perceptual quality across all planes. In contrast, SkelEM achieves the best perceptual quality in the XZ/YZ views while maintaining competitive fidelity scores across all metrics, demonstrating a favorable balance under zero-shot domain shift. This balance establishes a practical foundation for deploying pre-trained SkelEM to novel imaging devices.

\subsubsection{Segmentation Performance Comparison}
To assess the practical utility of our reconstructions, we evaluate 
downstream membrane segmentation performance on the EPFL test set. 
A public pre-trained segmentation model~\cite{zhang2024segneuron} is 
applied without fine-tuning to all reconstructed volumes, with results 
binarized via Otsu thresholding~\cite{otsu1975threshold}. Qualitative comparison is shown in \cref{figure:seg_fig}, where SkelEM produces 
membrane segmentations most consistent with those derived from the ground-truth volume. 
We report F1 score, Intersection over Union (IoU)~\cite{zhou2019iou}, Adapted Rand 
Error (ARE)~\cite{liu2014modular}, and Variation of Information 
(VoI)~\cite{meilua2007comparing}, using segmentation on the ground-truth volume as reference. As shown in \cref{tab:downstream_segmentation}, SkelEM achieves the 
best performance across all four metrics among self-supervised methods, 
even surpassing both supervised baselines. This is 
particularly notable given that SkelEM does not lead on PSNR or SSIM, 
demonstrating that pixel-wise fidelity metrics alone are insufficient 
proxies for structural quality in downstream analysis.

The ranking further validates our design choices. D2R-AENet, the 
strongest fidelity-oriented baseline, ranks second, confirming that 
structural fidelity contributes to segmentation accuracy. However, 
SkelEM's margin over D2R-AENet across all four metrics demonstrates 
that recovering realistic high-frequency membrane textures is critical 
for precise boundary delineation. The substantial drop from 
vEMDiffuse-i to vEMDiffuse-a further highlights the difficulty of 
maintaining structural accuracy under self-supervised constraints, which is
a challenge our training-signal decoupling directly addresses.

\subsection{Generalizability to Volume Light Microscopy}
\label{sec:vlm_generalizability}

To demonstrate generalization beyond VEM, we evaluate SkelEM on a 
real-world 10$\times$ ASR task using a public zebrafish retina VLM 
dataset \cite{weigert2018content}. Since isotropic ground truth is 
physically unattainable in standard VLM acquisition, quantitative 
evaluation is inherently infeasible in this setting, making visual 
assessment the standard practice for such real-world demonstrations 
\cite{lee2024reference, chen2025self}. As shown in 
\cref{figure:first_vis} (Bottom row, blue box), SkelEM recovers 
fine structural details from coarse anisotropic input, producing 
coherent 3D volumes without any supervised fine-tuning. This 
qualitative demonstration suggests that the training-signal decoupling 
design generalizes across fundamentally distinct imaging modalities.

% ============================================================
% Merged Ablation Table for SkelEM
% Combines: skeleton type, adaptation, and stage ablations
% ============================================================

\begin{table}[t]
\centering
\caption{Ablation study on the EPFL dataset ($r=8$). We analyze three design axes jointly:
(1) \textbf{Stage}: whether to use the full two-stage pipeline;
(2) \textbf{Skeleton}: the source of the initial structural prior;
(3) \textbf{Adapt}: whether to apply sparse data fine-tuning.
In the \textit{Refiner only} variant, the refiner receives the 
sparse observed slices $I_0, I_1$ directly as boundary conditions, 
with no structural skeleton prior provided as initialization.
}
\label{tab:ablation}
\setlength{\tabcolsep}{4pt}
\resizebox{\linewidth}{!}{%
\begin{tabular}{llc | c | c|c|c | c|c|c}
\toprule
\multirow{2}{*}{\textbf{Stage}} & \multirow{2}{*}{\textbf{Skeleton}} & \multirow{2}{*}{\textbf{Adapt}}
  & \multirow{2}{*}{3D-PSNR$\uparrow$}
  & \multicolumn{3}{c|}{SSIM$\uparrow$}
  & \multicolumn{3}{c}{LPIPS$\downarrow$} \\
\cmidrule{5-10}
& & & & XY & XZ & YZ & XY & XZ & YZ \\
\midrule
\multicolumn{10}{l}{\textit{(i) Necessity of the two-stage design}} \\
Skel only    & Ours & ---  & 25.96 & 0.6261 & 0.6627 & 0.6561 & 0.3466 & 0.5171 & 0.5210 \\
Refiner only    & --- & \texttimes  & 22.43 & 0.5007 & 0.5375 & 0.5308 & 0.3115 & 0.4390 & 0.4396 \\
Refiner only & ---  & \checkmark  & 20.26 & 0.4423 & 0.3533 & 0.3456 & 0.3075 & 0.5952 & 0.5980 \\
\midrule
\multicolumn{10}{l}{\textit{(ii) Quality of the skeleton prior}} \\
Two-stage & Bilinear             & \checkmark & 22.55 & 0.4822 & 0.4743 & 0.4635 & 0.3504 & 0.4071 & 0.4168 \\
Two-stage & RIFE (pretrained)    & \checkmark & 25.19 & 0.5973 & 0.6275 & 0.6216 & 0.3110 & 0.3542 & 0.3561 \\
Two-stage & InterpolAI (pretrained)                 & \checkmark & 24.78 & 0.5863 & 0.6057 & 0.6082 & 0.3354 & 0.3406 & 0.3419 \\
\midrule
\multicolumn{10}{l}{\textit{(iii) Effect of sparse data adaptation}} \\
Two-stage & Ours        & \texttimes               & 25.94 & 0.6372 & 0.6716 & 0.6659 & 0.4328 & 0.3976 & 0.4019 \\
Two-stage & Ours        & +Sparse3Diff~\cite{lee2025sparse3diff} & 25.59 & 0.6245 & 0.6578 & 0.6512 & 0.3388 & 0.3640 & 0.3701 \\
Two-stage & Ours        & \checkmark               & 25.56 & 0.6192 & 0.6532 & 0.6464 & 0.3049 & 0.3213 & 0.3298 \\
\bottomrule
\end{tabular}%
}
\vspace{-1.5em}
\end{table}

% ============================================================
% Merged Ablation Analysis — SkelEM (Section 4.4)
% ============================================================

\vspace{-1em}
\subsection{Ablation Study}
\label{sec:ablation}

We validate the three core design axes of SkelEM on the EPFL dataset ($r=8$),
with results reported across all three views in Tab.~\ref{tab:ablation}.

\noindent\textbf{Necessity of the two-stage design.}
The skeleton-only variant achieves the highest fidelity but suffers from severe axial blurring, confirming the over-smoothing tendency of convolutional priors. At the other extreme, the refiner-only variant without adaptation exhibits catastrophic fidelity collapse with severe axial artifacts, confirming that an unconstrained diffusion model cannot synthesize topologically consistent slices without structural guidance. Further applying $\mathcal{L}_{\mathrm{adapt}}$ to this skeleton-free refiner causes additional degradation, as corrupted anchors provide misleading alignment targets and destabilize the generative prior, which demonstrates that adaptation cannot substitute for a reliable structural anchor but rather acts as a refinement component. Only the full two-stage pipeline resolves this trilemma, maintaining competitive fidelity while delivering detailed axial reconstructions.

\noindent\textbf{Quality of the skeleton prior.}
Fixing the refinement stage and varying the skeleton source reveals a clear hierarchy. Bilinear interpolation, lacking structural priors, substantially degrades all metrics. While pre-trained VFI models provide useful motion priors, their training on natural video motion poorly generalizes to the nonlinear deformations characteristic of biological ultrastructures. Our skeleton, re-trained on the pseudo-HR volume $\hat{V}_{\mathrm{high}}$, achieves the best performance across all metrics, validating the necessity of domain-specific re-training.

\noindent\textbf{Effect of sparse data adaptation.}
Within the full two-stage framework, training without adaptation produces high fidelity but poor perceptual quality, as the refiner learns to reproduce the over-smoothed pseudo-HR textures. Substituting our strategy with Sparse3Diff~\cite{lee2025sparse3diff} partially recovers perceptual quality but remains inferior. Our $\mathcal{L}_{\mathrm{adapt}}$ achieves the best perceptual quality across all views with only a marginal fidelity cost, demonstrating that aligning the refiner to sparse real slices is essential for generating realistic high-frequency details.

% ============================================================
% Ablation: Initialization Prior (λ) × Refinement Steps (S)
% Clean table, timing comparison in caption
% ============================================================

\begin{table}[t]
\centering
\footnotesize
\caption{Ablation on residual injection strength ($\lambda$) and refinement steps~$S$.
$\lambda=1.0$ injects the full predicted residual $\Delta_\tau$ into the initial state;
$\lambda=0.0$ reduces to skeleton-only initialization without residual injection.
Per-patch inference times (256$\times$256, single RTX 4090) are listed per block.
For reference, other diffusion-based ASR methods require:
vEMDiffuse-i/a~\cite{lu2024diffusion} $\sim$2071\,ms,
Lee et al.~\cite{lee2024reference} $\sim$2558\,ms,
DiffuseIR~\cite{pan2023diffuseir} $\sim$8503\,ms,
TPDM~\cite{lee2023improving} $\sim$402788\,ms per patch (see Suppl. for details).}
\label{tab:abl_steps}
\setlength{\tabcolsep}{4pt}
\resizebox{\linewidth}{!}{%
\begin{tabular}{c | c | c | c | c|c|c | c|c|c}
\toprule
\multirow{2}{*}{Steps ($S$)} & \multirow{2}{*}{Time (ms)$\downarrow$} & \multirow{2}{*}{$\lambda$}
  & \multirow{2}{*}{3D-PSNR$\uparrow$}
  & \multicolumn{3}{c|}{SSIM$\uparrow$}
  & \multicolumn{3}{c}{LPIPS$\downarrow$} \\
  \cmidrule{5-10}
& & & & XY & XZ & YZ & XY & XZ & YZ \\
\midrule

\multirow{3}{*}{$S=1$} & \multirow{3}{*}{84}
& $1.00$ & 25.93 & 0.6348 & 0.6701 & 0.6636 & 0.3486 & 0.3969 & 0.4020 \\
& & $0.50$ & 25.11 & 0.5827 & 0.6206 & 0.6132 & 0.2759 & 0.2975 & 0.3058 \\
& & $0.00$ & 23.92 & 0.5113 & 0.5510 & 0.5428 & 0.2877 & 0.2997 & 0.3070 \\
\midrule

\multirow{3}{*}{$S=3$} & \multirow{3}{*}{151}
& $ 1.00$ & 25.56 & 0.6192 & 0.6532 & 0.6464 & 0.3049 & 0.3213 & 0.3298 \\
& & $0.50$ & 25.49 & 0.6166 & 0.6501 & 0.6433 & 0.3035 & 0.3172 & 0.3268 \\
& & $0.00$ & 25.51 & 0.6150 & 0.6484 & 0.6416 & 0.3034 & 0.3186 & 0.3280 \\
\midrule

\multirow{3}{*}{$S=5$} & \multirow{3}{*}{225}
& $1.00$ & 25.31 & 0.6107 & 0.6411 & 0.6344 & 0.3053 & 0.3275 & 0.3356 \\
& & $0.50$ & 25.33 & 0.6102 & 0.6406 & 0.6339 & 0.3056 & 0.3287 & 0.3366 \\
& & $0.00$ & 25.36 & 0.6093 & 0.6400 & 0.6334 & 0.3054 & 0.3304 & 0.3379 \\

\bottomrule
\end{tabular}%
}
\end{table}

\noindent\textbf{Impact of residual injection and refinement steps.}
We ablate the refinement steps $S \in \{1, 3, 5\}$ corresponding to
initiating the reverse diffusion at timesteps $t_S \in \{100, 300, 500\}$
of a 1000-step schedule, and the residual injection strength
$\lambda \in [0,1]$ governing our truncated sampling strategy:
\begin{equation}
\label{eq:lambda_mix}
x_{t_S} = \sqrt{\bar{\alpha}_{t_S}} \left( \hat{I}^{(skel)}_{\tau} + \lambda \cdot \Delta_\tau \right) + \sqrt{1 - \bar{\alpha}_{t_S}} \, \epsilon,
\end{equation}
where $\Delta_\tau$ is the residual predicted by $f_\Delta(\cdot)$ and
$\epsilon \sim \mathcal{N}(0,I)$ is Gaussian noise. Results are reported in Tab.~\ref{tab:abl_steps}.
At $S=1$, the full residual prior ($\lambda=1.0$) anchors the
single denoising step to a structurally informed starting point and achieves
peak fidelity, while reducing $\lambda$ to $0$ leads to severe structural collapse, 
indicating $\Delta_\tau$ provides essential structural information 
under minimal diffusion budget. Yet regardless of $\lambda$, $S=1$
yields poor perceptual quality, as one step cannot remove the regression bias
of the prior network.
At $S=3$, additional stochastic steps inject authentic
biological textures, sensitivity to $\lambda$ largely vanishes, and
$\lambda=1.0$ achieves the optimal equilibrium between structural fidelity and
perceptual realism. Further increasing to $S=5$ yields no quality gain while
slightly degrading fidelity, suggesting over-diffusion of the structural prior.
We therefore adopt $\lambda=1.0,\,S=3$ as our default.

\section{Conclusion}
\label{sec:conclusion}
In this work, we propose SkelEM, a novel two-stage self-supervised framework with training-signal decoupling between structural skeleton generation and skeleton-guided diffusion refinement. By identifying the absence of a domain-adapted structural skeleton as the shared root cause of over-smoothing and hallucination, SkelEM establishes a deterministic topological anchor that enables effective truncated diffusion in merely $\leq$5 steps. Extensive experiments demonstrate the most favorable balance across the fidelity-perception trade-off among self-supervised methods, with state-of-the-art downstream membrane segmentation performance and robust generalization across distinct imaging instruments and modalities, offering a practical solution for large-scale isotropic reconstruction in connectomics and cell biology.

\section*{Acknowledgements}
This work was supported by the project of high-throughput multifunctional scanning electron microscopy analytical system.

% ---- Bibliography ----
%
% BibTeX users should specify bibliography style 'splncs04'.
% References will then be sorted and formatted in the correct style.
%
\bibliographystyle{splncs04}
\bibliography{main}
% --- 引入补充材料 ---
\clearpage
\appendix % 如果需要自动切换为附录编号（如 A, B, C）

\section{Supplementary Materials Overview}

This supplementary material provides comprehensive theoretical analyses, implementation details, and extensive experimental evaluations to support the main manuscript. The document is organized as follows:

\begin{itemize}
    \item \cref{sec:supp_brave}: Details of our newly introduced BRAVE-ASR dataset used in cross-instrument validation, including imaging details, physical resolution, and other validating information. 
    \item \cref{sec:degradation}: Further, we provide validation of the physical realism of $r=8$ slice sampling strategy through theoretical noise modeling and empirical verification on the BRAVE-ASR benchmark.
    \item \cref{sec:implementation_details}: Details the algorithmic procedures for the multi-stage training and inference of SkelEM, accompanied by comprehensive qualitative comparisons on VEM datasets.
    \item \cref{sec:limitation}: Discusses the current limitations in resolving nano-scale geometric ambiguities, offering a transparent comparison with natural video priors.
\end{itemize}

\section{The BRAVE-ASR Dataset}
\label{sec:supp_brave}

\begin{figure}[tbp]
\centering
\includegraphics[width=1\textwidth]{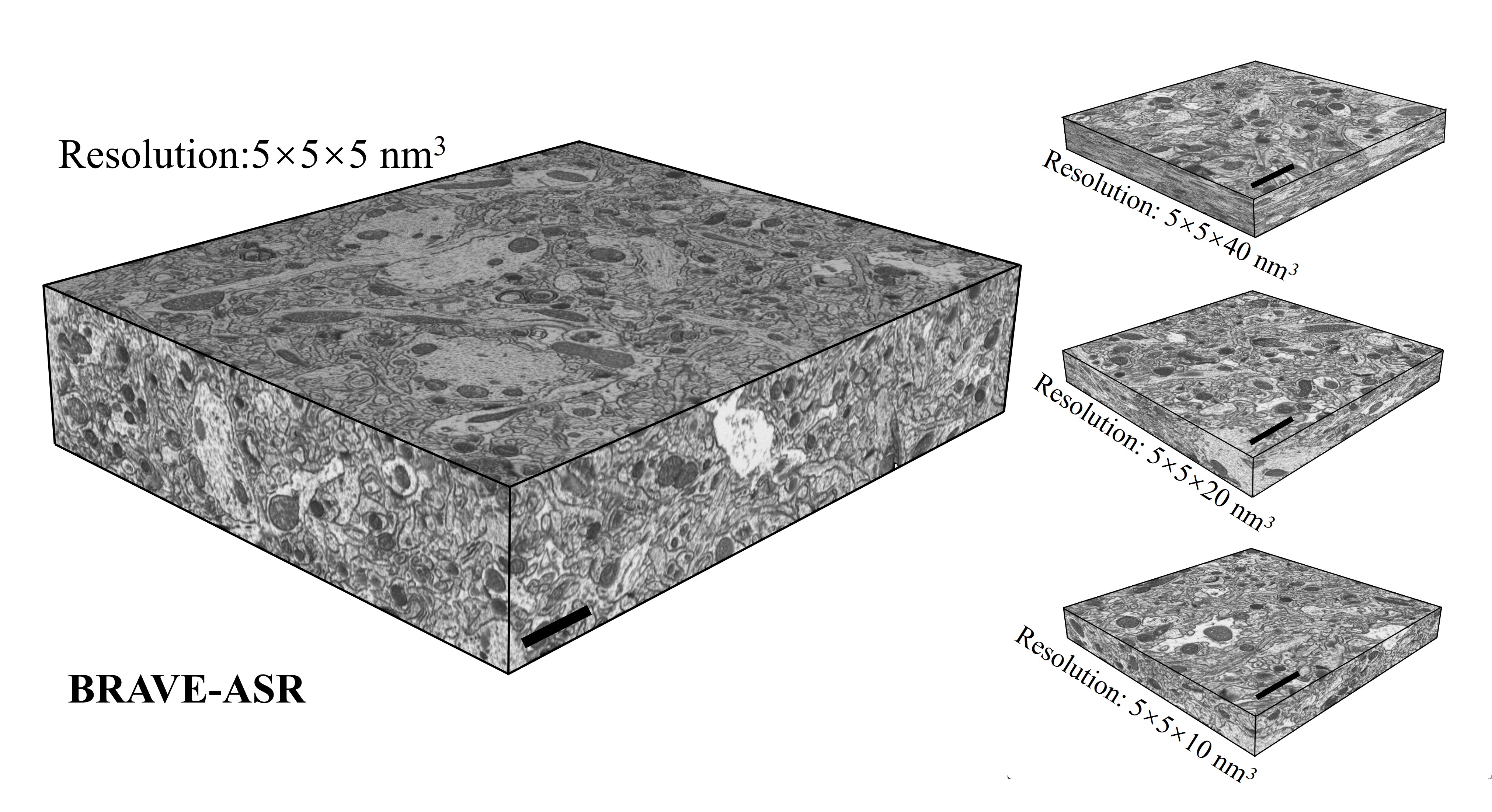}
\caption{Overview of the BRAVE-ASR dataset. Left: the isotropic reference volume ($3708 \times 3000 \times 800$ voxels at $5 \times 5 \times 5~\text{nm}^3$). Right: three co-aligned physically acquired anisotropic volumes ($2000 \times 2000 \times 256$ voxels each) at slice thicknesses of 10\,nm, 20\,nm, and 40\,nm, corresponding to $2\times$, $4\times$, and $8\times$ axial anisotropy. All volumes share the same 5\,nm lateral pixel size and were acquired from mouse brain tissue using a Thermo Scientific Helios Hydra Plasma FIB-SEM. Scale bars: 2\,$\mu$m.}
\label{fig:BRAVE-vis}
\end{figure}

As shown in \cref{fig:BRAVE-vis}, a contribution of our work is the \textbf{BRAVE-ASR} dataset 
(\textbf{B}enchma\textbf{R}king \textbf{A}nisotropic \textbf{V}olume-microscopy \textbf{E}valuation for \textbf{A}xial \textbf{S}uper-\textbf{R}esolution), 
which we acquired and have publicly released to establish a standardized benchmark for evaluating zero-shot instrument transfer in axial super-resolution methods. Besides, we use this dataset to prove that slice downsampling is the more physically faithful choice for ASR benchmarking compared to the average downsampling in \cref{sec:degradation}.

\subsection{Motivation}

Existing ASR benchmarks predominantly rely on FIB-SEM instruments (e.g., the Zeiss NVision40 used for the EPFL dataset), making it impossible to evaluate whether methods generalize across fundamentally different imaging platforms.
BRAVE-ASR addresses this gap by providing co-aligned volumes acquired at multiple physical slice thicknesses on a fundamentally different instrument, enabling both (1) rigorous validation of degradation simulation strategies against physically acquired ground truth, and (2) standardized evaluation of cross-instrument generalization.

\subsection{Acquisition and Specifications}

All four volumes were acquired from the \emph{same} mouse brain tissue block using a different FIB-SEM microscope (Thermo Scientific Helios Hydra Plasma FIB-SEM), with a fixed lateral pixel size of 5\,nm and a dwell time of 1\,$\mu$s per pixel. This instrument differs substantially from the Zeiss NVision40 Gallium FIB-SEM used for EPFL, exhibiting distinct noise characteristics, contrast profiles, and milling artifacts due to the use of a xenon plasma ion source. By imaging all volumes from a single tissue block under identical beam conditions, we ensure that the only controlled variables across the four volumes are the axial sectioning thickness and the spatial position within the block.

The dataset comprises the following co-aligned volumes:
\begin{itemize}
    \item \textbf{Isotropic reference volume}: $3708 \times 3000 \times 800$ voxels (height $\times$ width $\times$ depth) at $5 \times 5 \times 5$\,nm$^3$ resolution. This serves as the high-resolution ground truth and the source for simulated degradation.
    \item \textbf{Physically acquired anisotropic volumes}: Three volumes at physical slice thicknesses of 10\,nm ($2\times$), 20\,nm ($4\times$), and 40\,nm ($8\times$), each with a base block size of $2000 \times 2000 \times 256$ voxels. All volumes share the same 5\,nm lateral pixel size and dwell time as the isotropic reference, differing only in axial sectioning thickness.
\end{itemize}

Since the four volumes originate from the same tissue block, they share identical biological content, staining, and embedding conditions. All volumes have undergone SIFT-based affine registration for spatial co-alignment and histogram normalization to account for minor intensity variations across imaging sessions. The registration quality was manually verified by visual inspection, confirming that the co-aligned volumes can be regarded as well-registered blocks whose primary difference is the physical slice thickness.

\subsection{Usage Protocol}

In our paper, BRAVE-ASR is used exclusively as a \textbf{zero-shot transfer test set}. All models are trained on the EPFL dataset (with Zeiss NVision40 FIB-SEM, mouse hippocampus, 5\,nm isotropic) and evaluated directly on BRAVE-ASR without any fine-tuning. This protocol quantifies the degree to which each method's learned representations are instrument-specific versus genuinely capturing biological structure priors.

In addition, the physically acquired multi-thickness volumes enable the degradation model validation presented in \cref{sec:degradation}: by comparing simulated anisotropic volumes (via slice sampling or average downsampling of the 5\,nm reference) against the corresponding physically acquired volumes at 10\,nm, 20\,nm, and 40\,nm, we can directly assess which simulation strategy better preserves real acquisition statistics (see \cref{fig:different_strategy} and \cref{tab:fid_degradation}).

The BRAVE-ASR dataset has been publicly released on Zenodo at \url{https://doi.org/10.5281/zenodo.18920195}. We encourage the community to adopt BRAVE-ASR as a standardized benchmark for cross-instrument evaluation and degradation benchmark in axial super-resolution research.

\section{Analysis of the Degradation Model}
\label{sec:degradation}

In the paper, we adopted an $r=8$ axial slice sampling strategy to simulate anisotropic volumes, aligned with downsampling method used in \cite{lu2024diffusion}. This is a common and critical choice in ASR research, as the choice of degradation model directly impacts the physical realism of the benchmark. In this section, we provide a detailed justification for our choice by first presenting a theoretical noise analysis and then a rigorous experimental validation on BRAVE-ASR dataset.

\subsection{Theoretical Noise Modeling in Volume Microscopy}
\label{sec:noise_modeel}

Taking into account a real-world volume microscopy slice $y$, we can represent the microscopy image as the sum of a noise-free content term $x$ and a Poisson-Gaussian noise term $\epsilon_{PG}$. As described in \cite{foi2008practical}, the noise introduced during the imaging process can be expressed as:
$\epsilon_{PG} = \alpha \epsilon_P + \epsilon_G$ , where $\epsilon_P \sim \text{Poisson}(x / \alpha)$ and $\epsilon_G \sim \mathcal{N}(0, \sigma^2)$, with $\alpha$ being a process-dependent factor. In \cite{hasinoff2021photon}, the authors showed that Poisson-Gaussian noise $\epsilon_{PG}$ can be closely approximated as signal-dependent Gaussian noise: $\epsilon_G' \sim \mathcal{N}(0, \alpha x + \sigma^2)$. This simplifies the noisy volume microscopy slice $y$ to:
\begin{equation}
    y = x + \epsilon_G' , \ \epsilon_G' \sim \mathcal{N}(0, \alpha x + \sigma^2) .
    \label{equ:base_noise}
\end{equation}
We treat \cref{equ:base_noise} as the basic noise model for a single high-resolution slice.

\subsection{Comparison of Degradation Strategies}
\label{sec:theoritical_analysis}
\subsubsection{Average down-sampling:}
This common strategy generates an anisotropic slice $\hat{y}$ by averaging $r$ slices $\{y_i\}_{i=1}^r$:
\begin{equation}
    \label{equ:infos}
    \hat{y} = \frac{1}{r} \sum_{i=1}^r y_i = \frac{1}{r} \sum_{i=1}^r x_i + \frac{1}{r} \sum_{i=1}^r \epsilon_i = \hat{x} + \hat{\epsilon},
\end{equation}
 where \( \hat{x} = \frac{1}{r} \sum_{i=1}^r x_i \) and \( \hat{\epsilon} = \frac{1}{r} \sum_{i=1}^r \epsilon_i \). Based on the properties of independent Gaussian random variables, where the sum of $r$ variables with variance $\sigma$ has a variance of $r\times\sigma$, the noise term $\hat{\epsilon}$ follows:
\begin{equation}
    \hat{\epsilon} \sim \mathcal{N} \left( 0, \frac{1}{r^2} \sum_{i=1}^r (\alpha x_i + \sigma^2) \right) = \mathcal{N} \left( 0, \frac{1}{r} (\alpha \hat{x} + \sigma^2) \right).
    \label{equ:downsample_noise}
\end{equation}

As a result, the slice $\hat{y}$ synthesized by average-downsampling is
\begin{equation}
    \hat{y} = \hat{x} + \hat{\epsilon}, \ \hat{\epsilon} \sim \mathcal{N} \left( 0, \frac{1}{r} (\alpha \hat{x} + \sigma^2) \right).
    \label{equ:comparision_between}
\end{equation}

As shown in \cref{equ:comparision_between}, it is evident that the average down-sampling strategy artificially reduces the noise variance in the synthesized low-resolution slice $\hat{y}$ by a factor of $1/r$. Besides, it explicitly incorporates information from all layers $\{x_i\}_{i=1}^r$ to the input layer $\hat{y}$, as \cref{equ:infos} shows.

\subsubsection{Slice Sampling:}
In our experiments, we adopt the data preparation method from \cite{lu2024diffusion}, which simulates the anisotropic volume by retaining only a subset of slices. Starting from slice $y_0$, only the slices with indices $i$ such that $i \pmod r = 0$ are kept, resulting in a sequence $\{y_i\}_{i\pmod r =0}$. This slice sampling strategy does not average out the noise. Each retained slice in $\{y_i\}$ is a true slice from the original high-resolution volume and thus fully preserves the original, realistic noise distribution described in \cref{equ:base_noise}.

As a result, the slice sampling strategy provides a more challenging and physically faithful benchmark, as the ASR model must \textit{simultaneously} interpolate the missing $(r-1)$ slices \textit{and} handle the true, un-attenuated noise level of the acquisition system.

\begin{figure}[!tbp]
    \centering
    \begin{subfigure}{0.48\linewidth}
        \centering
        \resizebox{\columnwidth}{!}{%
        \includegraphics{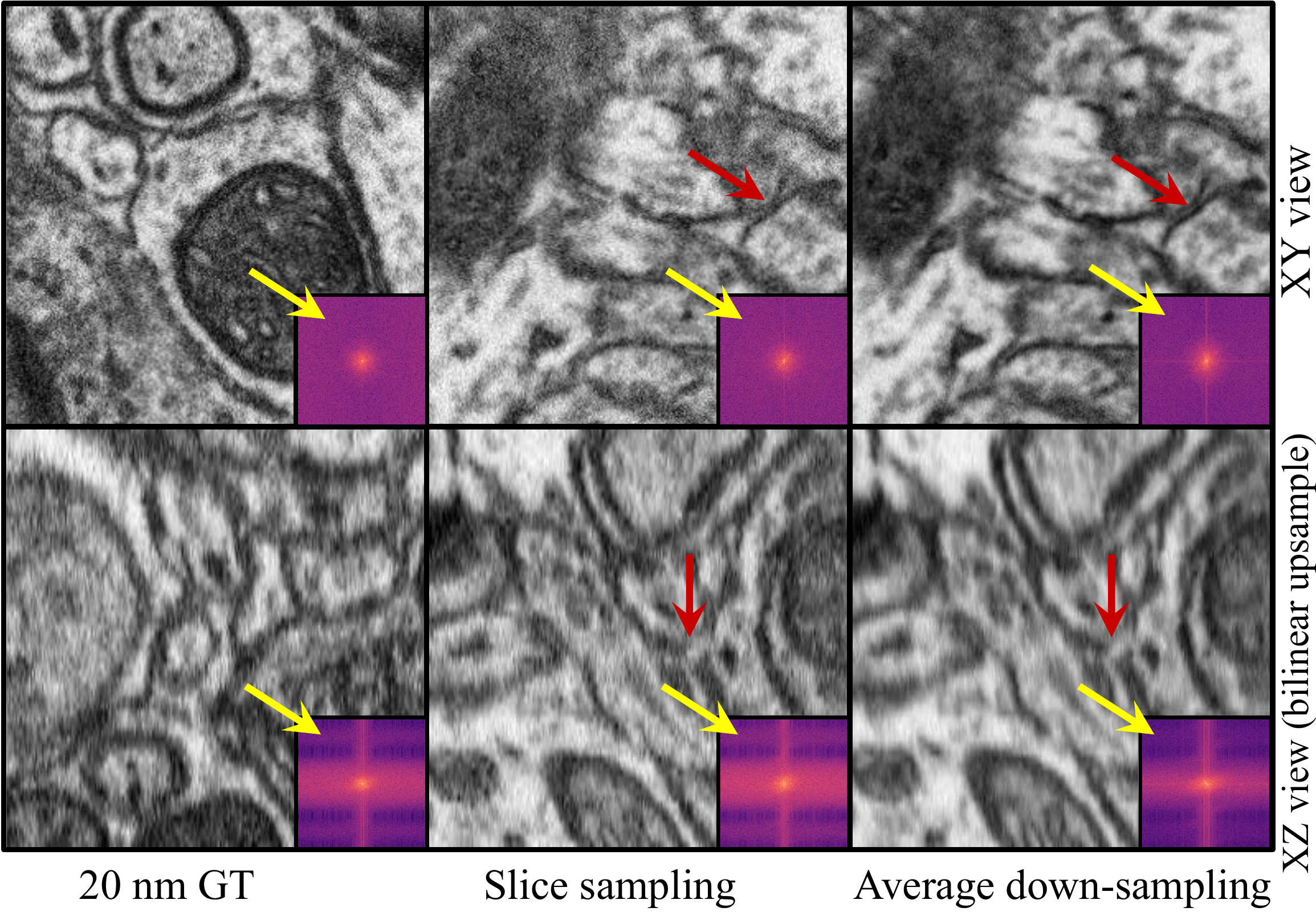}
        }
    \end{subfigure}
    \begin{subfigure}{0.48\linewidth}
        \centering
        \resizebox{\columnwidth}{!}{%
        \includegraphics{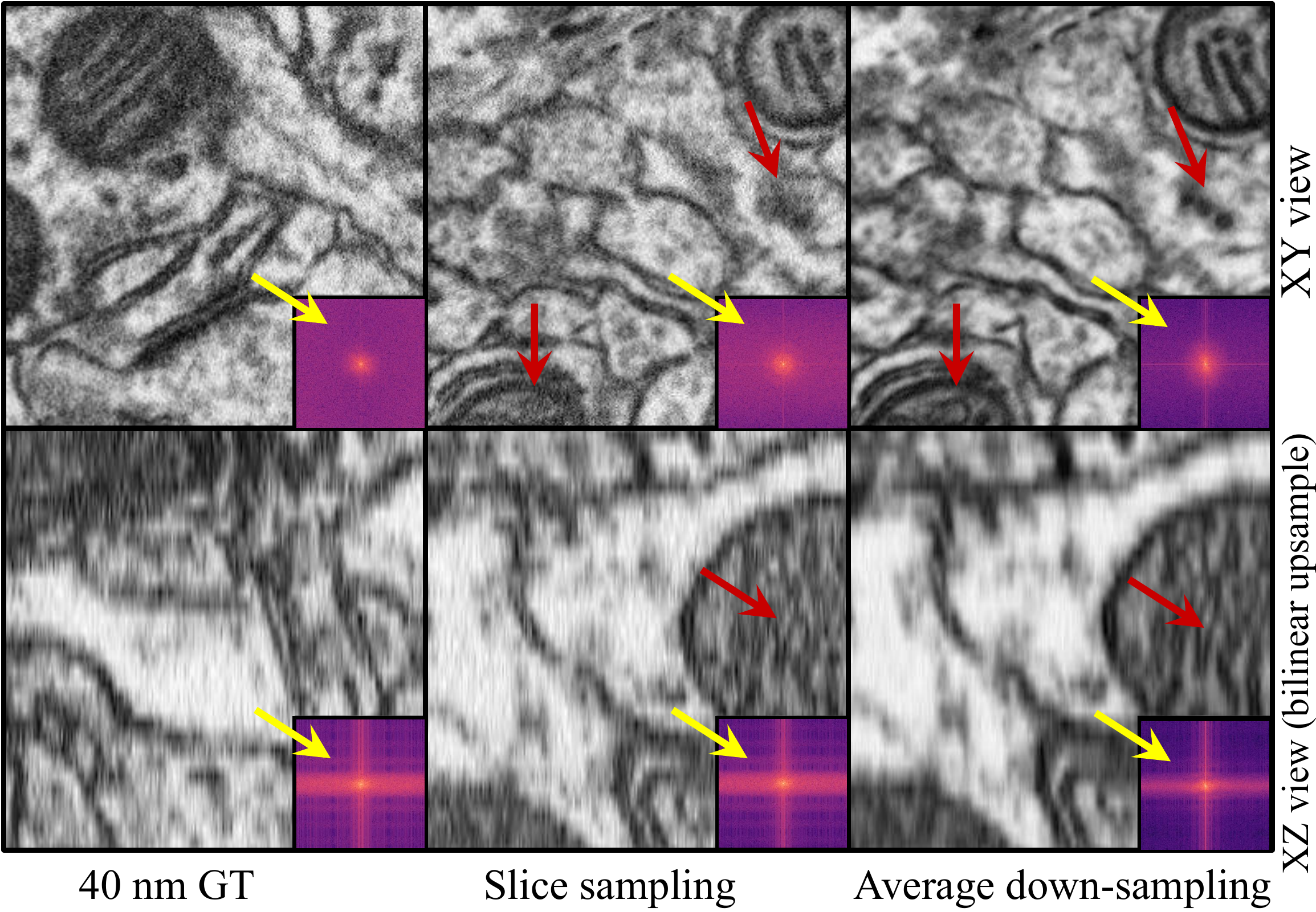}
        }
    \end{subfigure}
    \caption{\textbf{Qualitative comparison of degradation strategies on BRAVE-ASR at 20\,nm (left) and 40\,nm (right) slice thicknesses.} Each panel shows the XY (upper row) and XZ (lower row, bilinear upsampled for display) views for real GT, slice sampling, and average downsampling. Red arrows: ultrastructural details; yellow arrows: noise suppression. Inset Fourier spectra confirm that average downsampling severely attenuates high frequencies, while slice sampling faithfully preserves the spectral profile of real acquisitions.}
    \label{fig:different_strategy}
\end{figure}

\begin{figure}[tbp]
\centering
\includegraphics[width=0.6\textwidth]{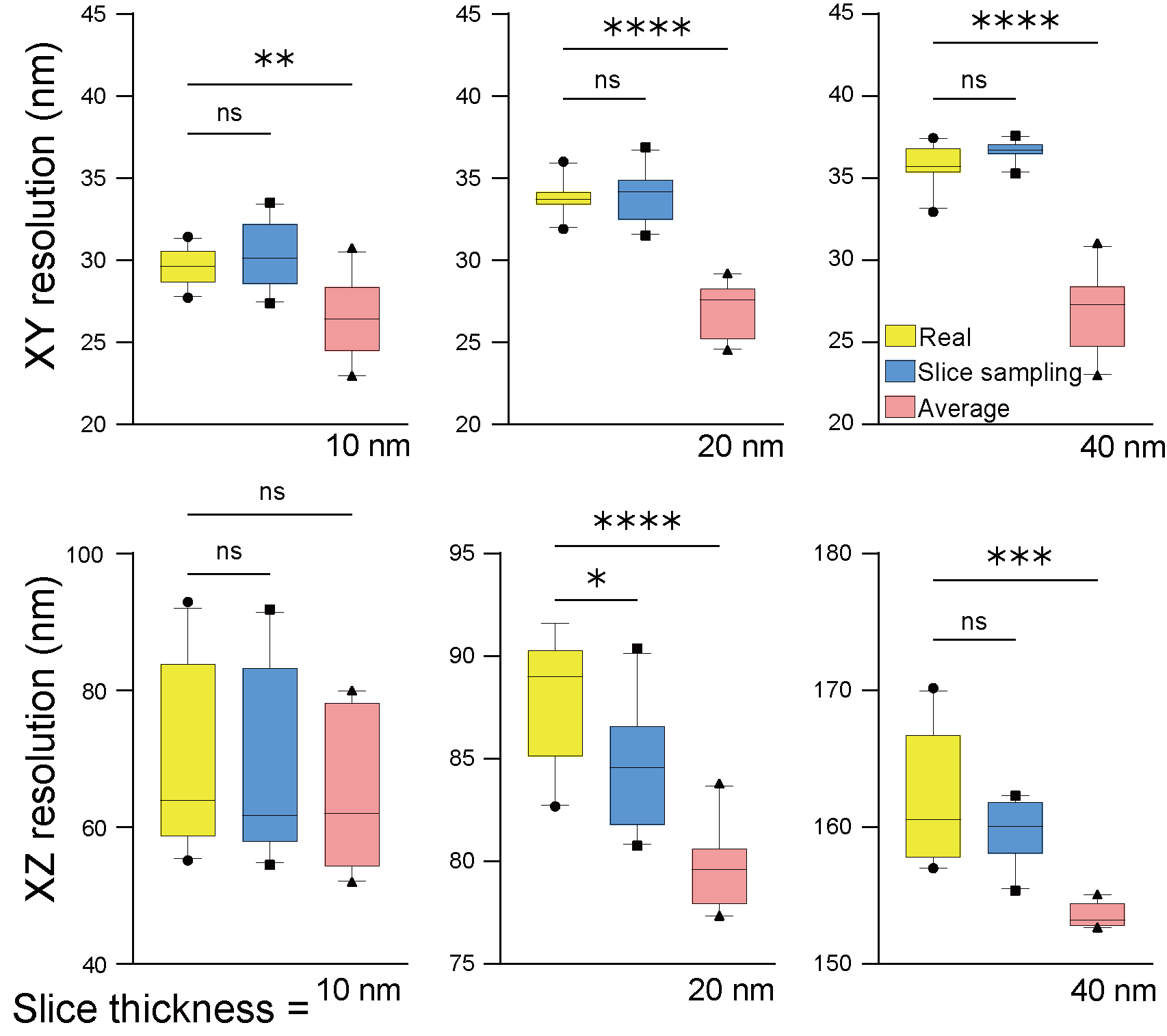}
\caption{\textbf{Quantitative Resolution Comparison on BRAVE-ASR.} 
Effective resolution in XY and XZ planes for slice sampling (blue), 
average downsampling (pink), and physically acquired data (yellow) 
at 10\,nm ($2\times$), 20\,nm ($4\times$), and 40\,nm ($8\times$) 
slice thicknesses. Statistical significance is indicated 
(ns: not significant; *: $p<0.05$; **: $p<0.01$; 
***: $p<0.001$; ****: $p<0.0001$).}
\label{fig:resotest}
\end{figure}

\subsection{Experimental Validation on the BRAVE-ASR Dataset}

To compare the noise and frequency characteristics, we quantify the effective volume resolution of the generated low-resolution inputs versus the physically acquired data by sectional Fourier Shell Correlation (SFSC)\cite{miplib2019}, which has been widely used in both VEM\cite{he2023isovem, lu2024diffusion} and VLM\cite{koho2019fourier}. Additionally, we compute the Fréchet Inception Distance 
(FID)\cite{heusel2017gans} between the simulated and physically acquired volumes to evaluate the distributional similarity in learned feature space, complementing the frequency-domain analysis of SFSC with a perceptual-level assessment.

\paragraph{Qualitative comparison.}
We first provide visual evidence in \cref{fig:different_strategy}, which compares volumes generated by slice sampling and average downsampling against the physically acquired ground truth at 20\,nm ($4\times$) and 40\,nm ($8\times$) thicknesses, in both the lateral (XY) and axial (XZ) views. Two trends are immediately apparent. First, the average downsampling strategy noticeably suppresses noise and fine-grained texture (yellow arrows), producing images that appear artificially clean compared to the real acquisitions. Second, it attenuates ultrastructural details (red arrows) that are well preserved by slice sampling. The corresponding 2D Fourier spectra confirm this observation: average downsampling yields a visibly contracted spectral distribution, indicating significant loss of high-frequency content, whereas slice sampling retains a spectral profile closely matching the ground truth.

\paragraph{Frequency-domain resolution analysis.}
To quantify these observations, we measure the effective volumetric resolution using sectional Fourier Shell Correlation (SFSC)~\cite{miplib2019}, a standard metric widely adopted in both VEM~\cite{he2023isovem, lu2024diffusion} and VLM~\cite{koho2019fourier}. As shown in \cref{fig:resotest}, for all tested slice thicknesses (10\,nm, 20\,nm, and 40\,nm, corresponding to $2\times$, $4\times$, and $8\times$ axial downsampling), the XY resolution of slice-sampled volumes shows no statistically significant difference from the physically acquired data. In contrast, the average downsampling method causes a highly significant reduction in XY resolution at the $4\times$ and $8\times$ settings, confirming that it over-smooths the original high-frequency information. In the XZ plane, while minor deviations begin to appear at higher downsampling factors for both strategies, slice sampling remains substantially closer to the real data than average downsampling, which introduces statistically significant resolution loss.

\paragraph{Distributional similarity analysis.}
To complement the frequency-domain analysis, we compute the Fr\'{e}chet Inception Distance (FID)~\cite{heusel2017gans} between simulated and physically acquired volumes. As shown in \cref{tab:fid_degradation}, slice sampling achieves substantially lower FID across all planes and downsampling factors. The gap widens with the downsampling factor: at $8\times$, slice sampling yields an FID of 8.64 versus 116.52 for average downsampling in the XY plane. These results confirm that average downsampling introduces a scale-dependent distributional shift, consistent with the theoretical analysis in \cref{sec:theoritical_analysis}.

We acknowledge that neither simulation strategy can fully replicate the complex physics of real volumetric acquisition, where factors such as charging effects, beam damage, and depth-dependent contrast variations introduce degradations beyond simple noise modeling. Nevertheless, the above qualitative, frequency-domain, and distributional analyses provide converging evidence that, among the two widely adopted simulation strategies, slice sampling is the more physically faithful choice for ASR benchmarking. In our work, we adopt $r=8$ as our primary benchmark setting, consistent with prior work\cite{chen2025self, he2023isovem, lee2024reference}, and validated by BRAVE-ASR as the most physically faithful simulation under this factor. It better preserves the native noise statistics and high-frequency structural content of real acquisitions, whereas average downsampling introduces artificial noise suppression and spectral attenuation that do not reflect real-world imaging conditions.

\begin{table}[t]
\centering
\caption{\textbf{Fr\'{e}chet Inception Distance (FID) between simulated and physically acquired anisotropic volumes on BRAVE-ASR.} Lower FID indicates greater distributional similarity to real acquisitions. FID is computed from 5{,}000 randomly sampled patches per condition. Slice sampling consistently achieves substantially lower FID than average downsampling across all planes and downsampling factors.}
\begin{tabular}{cc|c|c}
\toprule
Scale & Plane & Slice Sampling ($\downarrow$) & Avg. Downsampling ($\downarrow$) \\
\midrule
\multirow{3}{*}{$2\times$ (10\,nm)} 
 & XY & \textbf{22.43} & 61.96 \\
 & XZ & \textbf{17.29} & 44.66 \\
 & YZ & \textbf{15.05} & 42.74 \\
\midrule
\multirow{3}{*}{$4\times$ (20\,nm)} 
 & XY & \textbf{12.42} & 91.34 \\
 & XZ & \textbf{14.37} & 60.96 \\
 & YZ & \textbf{8.99} & 45.20 \\
\midrule
\multirow{3}{*}{$8\times$ (40\,nm)} 
 & XY & \textbf{8.64} & 116.52 \\
 & XZ & \textbf{5.29} & 54.11 \\
 & YZ & \textbf{4.72} & 54.21 \\
\bottomrule
\end{tabular}%
\label{tab:fid_degradation}
\end{table}

\section{SkelEM Implementation Details}
\label{sec:implementation_details}

\subsection{Pseudo Code for Training and Inference}

The SkelEM framework involves three training stages and a fast inference procedure, following the order illustrated in Fig.~2 of the main paper. We detail: (1)~training the topological skeleton network on synthetic manifolds in \cref{alg:skeleton_training}, (2)~cycle-consistent prior extraction on sparse real slices in \cref{alg:prior_extraction}, (3)~diffusion refiner base training and sparse self-alignment in \cref{alg:refiner_training}, and (4)~the prior-truncated inference process in \cref{alg:inference_final}.

%% ============ Algorithm 1: Skeleton Training ============
\begin{algorithm}[!t]
\caption{Stage 1: Topological Skeleton via Synthetic Manifold Pre-training}
\begin{algorithmic}[1]
\REQUIRE Pseudo-HR volume $\hat{V}_{high}$; low resolution volume $V_{low}$; ASR upsampling factor $r$.
\STATE Initialize flow network $f_{flow}$ (with detail refiner $f_{detail}$).
\FOR{epochs}
    \STATE Sample adjacent slices $(I_0, I_1)$ from $V_{low}$ and target $I_\tau$ from $\hat{V}_{high}$, where $\tau \in (0,1)$.
    \STATE Predict bidirectional flows $F_{\tau \rightarrow 0}$, $F_{\tau \rightarrow 1}$ and blending mask $M_\tau$ via $f_{flow}$.
    \STATE Synthesize skeleton (Eq.~1 in main paper):
    $$\hat{I}^{(skel)}_{\tau} = M_\tau \odot \mathcal{W}(I_0, F_{\tau \rightarrow 0}) + (1 - M_\tau) \odot \mathcal{W}(I_1, F_{\tau \rightarrow 1})$$
    \STATE Minimize $\mathcal{L}_{1}(\hat{I}^{(skel)}_{\tau}, I_\tau)$.
\ENDFOR
\STATE \textbf{Discard} $f_{detail}$ to enforce topology--texture separation.
\STATE \textbf{Freeze} $f_{flow}$.
\ENSURE Frozen topological skeleton network $f_{flow}$.
\end{algorithmic}
\label{alg:skeleton_training}
\end{algorithm}

%% ============ Algorithm 2: Cycle-Consistent Prior Extraction ============
\begin{algorithm}[!t]
\caption{Stage 2: Cycle-Consistent Prior Extraction}
\begin{algorithmic}[1]
\REQUIRE Frozen skeleton network $f_{flow}$; low resolution volume $V_{low}$.
\STATE Initialize residual estimator $f_\Delta$.
\FOR{epochs}
    \STATE Sample three consecutive slices $\{I_{-1}, I_0, I_1\}$ from $V_{low}$.
    \STATE Sample relative position $\tau \in (0,1)$.
    \STATE \textbf{// Cyclic reconstruction of observed slice $I_0$}
    \STATE Warp virtual anchor $I_{-\tau}$ from $(I_{-1}, I_0)$ and $I_{1-\tau}$ from $(I_0, I_1)$ using frozen $f_{flow}$.
    \STATE Cyclically reconstruct (Eq.~4 in main paper):
    $$\tilde{I}_0 = \mathcal{W}_{cycle}(I_{-\tau},\; I_{1-\tau},\; f_{flow},\; \tau)$$
    \STATE Compute ground-truth residual: $\Delta_{gt} = I_0 - \tilde{I}_0$.
    \STATE \textbf{// Train residual estimator via frequency loss}
    \STATE Compute skeleton at observed position: $\hat{I}^{(skel)}_{\tau}$ from $f_{flow}$.
    \STATE Minimize $\mathcal{L}_{freq}$ (Eq.~5 in main paper):
    $$\mathcal{L}_{freq} = \| \mathcal{F}_{fft}(f_{\Delta}(\hat{I}^{(skel)}_{\tau})) - \mathcal{F}_{fft}(\Delta_{gt}) \|_1$$
\ENDFOR
\ENSURE Trained residual estimator $f_\Delta$.
\end{algorithmic}
\label{alg:prior_extraction}
\end{algorithm}

%% ============ Algorithm 3: Diffusion Base Training + Adaptation ============
\begin{algorithm}[!t]
\caption{Stage 3: Diffusion Base Training and Sparse Self-Alignment}
\begin{algorithmic}[1]
\REQUIRE Pseudo-HR volume $\hat{V}_{high}$; low resolution volume $V_{low}$; upsampling factor $r$.

\STATE \textbf{// Phase A: Base training on synthetic manifolds}
\STATE Initialize diffusion refiner $f_\theta$.
\FOR{epochs}
    \STATE Sample target $\hat{I}_\tau$ from $\hat{V}_{high}$ with boundary slices $I_0, I_1$ from $V_{low}$.
    \STATE Sample noise $\epsilon \sim \mathcal{N}(0, \mathbf{I})$ and timestep $t$.
    \STATE Minimize $\mathcal{L}_{base}$ (Eq.~2 in main paper):
    $$\mathcal{L}_{base} = \mathbb{E}_{t, \hat{I}_\tau, \epsilon} \left[ \| \epsilon - f_{\theta}(x_{t}, t, I_0, I_1, \tau) \|_2^2 \right]$$
\ENDFOR

\STATE
\STATE \textbf{// Phase B: Sparse self-alignment on real slices}
\FOR{epochs}
    \STATE Sample three consecutive slices $\{I_{-1}, I_0, I_1\}$ from $V_{low}$.
    \STATE Sample relative position $\tau \in (0,1)$.
    \STATE Generate intermediate anchors using current $f_\theta$:
    $$\hat{I}_{\tau-1} = \text{Sample}(f_{\theta},\; \text{cond}=(I_{-1}, I_0),\; \text{pos}=\tau)$$
    $$\hat{I}_{\tau} = \text{Sample}(f_{\theta},\; \text{cond}=(I_{0}, I_1),\; \text{pos}=\tau)$$
    \STATE Construct bidirectional conditioning set:
    $$\Phi = \left\{(\hat{I}_{\tau-1},\; \hat{I}_{\tau},\; 1{-}\tau),\;\; (\hat{I}_{\tau},\; \hat{I}_{\tau-1},\; \tau)\right\}$$
    \STATE Minimize $\mathcal{L}_{adapt}$ (Eq.~3 in main paper), with $\hat{I}_{\tau-1}, \hat{I}_{\tau}$ \textbf{detached} (no gradient):
    $$\mathcal{L}_{adapt} = \sum_{\phi \in \Phi} \mathbb{E}_{\epsilon, I_0, t} \left[ \| \epsilon - f_{\theta}(x_{t}, t, \phi) \|_2^2 \right]$$
\ENDFOR
\ENSURE Adapted refiner $f_\theta$.
\end{algorithmic}
\label{alg:refiner_training}
\end{algorithm}

%% ============ Algorithm 4: Inference ============
\begin{algorithm}[!t]
\caption{Prior-Truncated Axial Super-Resolution Inference}
\begin{algorithmic}[1]
\REQUIRE Low resolution volume $V_{low}$; upsampling factor $r$; refinement steps $S$ ($S \leq 5$); adapted refiner $f_\theta$; frozen skeleton network $f_{flow}$; residual estimator $f_\Delta$.
\STATE Initialize $V_{high} = V_{low}$.
\FOR{each pair of adjacent slices $(I_0, I_1)$ in $V_{low}$}
    \FOR{target position $\tau = \frac{1}{r},\; \frac{2}{r},\; \ldots,\; \frac{r-1}{r}$}
        \STATE \textbf{// Deterministic skeleton generation}
        \STATE $\hat{I}^{(skel)}_{\tau} = f_{flow}(I_0, I_1, \tau)$ \hfill $\triangleright$ Eq.~1
        \STATE \textbf{// Cycle-consistent residual prediction}
        \STATE $\Delta_\tau = f_{\Delta}(\hat{I}^{(skel)}_{\tau})$
        \STATE \textbf{// Prior-truncated initialization}
        \STATE Aggregate prior: $\hat{I}^{(prior)}_{\tau} = \hat{I}^{(skel)}_{\tau} + \Delta_\tau$
        \STATE Sample $\epsilon \sim \mathcal{N}(0, \mathbf{I})$.
        \STATE $x_{t_S} = \sqrt{\bar{\alpha}_{t_S}} \, \hat{I}^{(prior)}_{\tau} + \sqrt{1 - \bar{\alpha}_{t_S}} \, \epsilon$ \hfill $\triangleright$ Eq.~6
        \STATE \textbf{// Rapid iterative refinement ($S$ steps)}
        \FOR{$n = S$ down to $1$}
            \STATE Predict clean state (Eq.~7 in main paper):
            $$\hat{x}_0 = \frac{1}{\sqrt{\bar{\alpha}_{t_n}}} \left( x_{t_n} - \sqrt{1 - \bar{\alpha}_{t_n}} \, f_{\theta}(x_{t_n}, t_n, I_0, I_1, \tau) \right)$$
            \STATE DDPM reverse step (Eq.~8 in main paper):
            $$x_{t_{n-1}} = \frac{\sqrt{\alpha_{t_n}}(1 - \bar{\alpha}_{t_{n-1}})}{1 - \bar{\alpha}_{t_n}} x_{t_n} + \frac{\sqrt{\bar{\alpha}_{t_{n-1}}}(1 - \alpha_{t_n})}{1 - \bar{\alpha}_{t_n}} \hat{x}_0 + \sigma_{t_n} z$$
        \ENDFOR
        \STATE Insert refined slice $\hat{I}_\tau = x_{t_0}$ into $V_{high}$.
    \ENDFOR
\ENDFOR
\ENSURE High-resolution volume $V_{high}$.
\end{algorithmic}
\label{alg:inference_final}
\end{algorithm}

\subsection{Qualitative Results on FIB-25 and EPFL}

\begin{figure*}[tbp]
\centering
\includegraphics[width=1\textwidth]{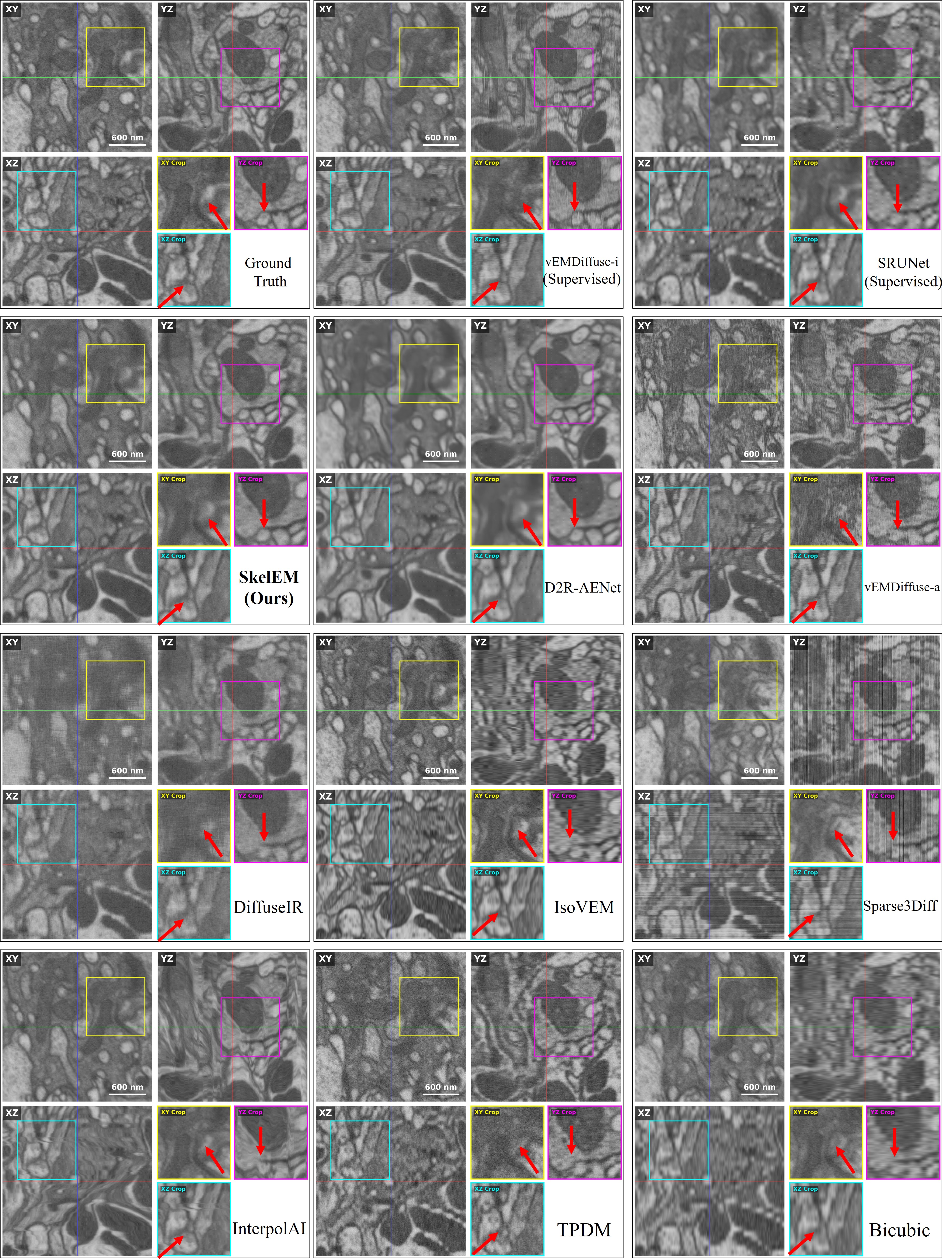}
\caption{Tri-view (XY, XZ, YZ) qualitative comparison on the FIB-25 dataset ($8\times$ ASR). Color-coded insets and red arrows highlight structural differences. Supervised methods are marked with (Sup.). $\tau$ of XY slice is $0.5$.}
\label{fig:FIB25_show}
\end{figure*}

\begin{figure*}[tbp]
\centering
\includegraphics[width=1\textwidth]{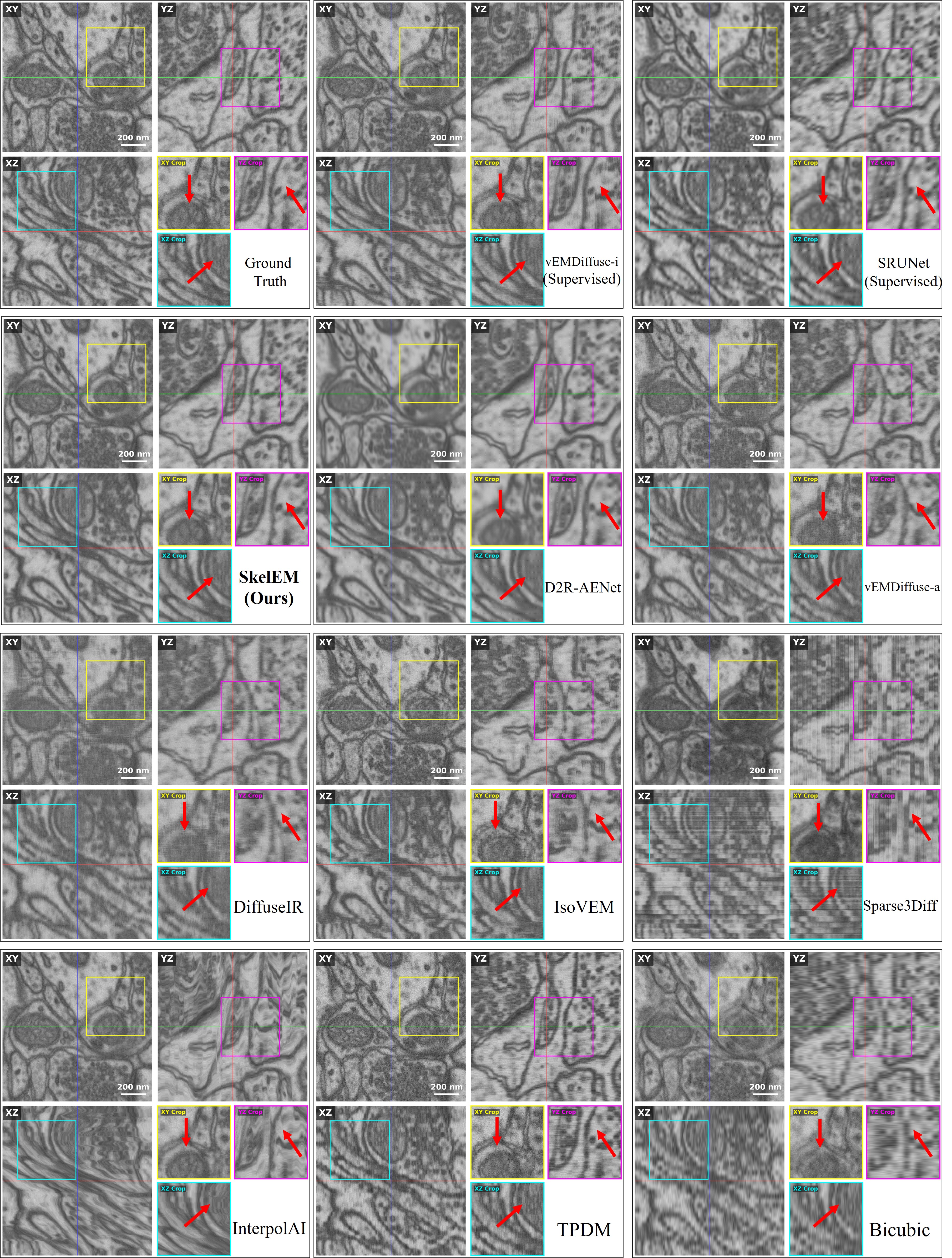}
\caption{Tri-view (XY, XZ, YZ) qualitative comparison on the EPFL dataset ($8\times$ ASR). Color-coded insets and red arrows highlight structural differences. Supervised methods are marked with (Sup.). $\tau$ of XY slice is $0.5$.}
\label{fig:EPFL_show}
\end{figure*}

While Fig.~1 in the main paper highlights performance on zero-shot instrument transfer and real-world scenarios, here we extend our qualitative evaluation to standard simulated anisotropic benchmarks. \cref{fig:FIB25_show} and \cref{fig:EPFL_show} provides a comprehensive tri-view (XY, XZ, YZ) comparison of SkelEM against state-of-the-art supervised and self-supervised ASR methods on the FIB-25 and EPFL datasets ($r=8$), with isotropic ground truth as reference. For each method, we display the three orthogonal views alongside color-coded cropped insets to facilitate detailed inspection of structural fidelity. Red arrows indicate regions of particular interest where methods diverge.

The visual comparisons reveal distinct failure modes across existing paradigms. Regression-based methods, including the supervised SRUNet and the self-supervised D2R-AENet, produce over-smoothed results where high-frequency textures such as membrane boundaries are lost. Video frame interpolation methods (InterpolAI) may appear plausible in the XY view but reveal severe structural discontinuities in the axial views (XZ, YZ) due to the domain gap between natural video and biological tissue. Diffusion-based methods like DiffuseIR frequently hallucinate structures in the XY plane, while TPDM shows significant discrepancies in detail recovery.

In contrast, SkelEM achieves a compelling balance across all three orthogonal views. The XY view preserves realistic biological textures with sharp membrane delineation; the XZ and YZ views demonstrate smooth axial continuity without the banding artifacts or structural hallucinations that plague other methods. Crucially, our self-supervised method recovers fine details comparable to the supervised generative baseline, as evidenced by the cropped insets, demonstrating that training-signal decoupling between topology and detail enables high-fidelity reconstruction without requiring paired isotropic training data.

\subsection{Overall Performance Balance}

\begin{figure}[!h]
  \centering
  \includegraphics[width=1\linewidth]{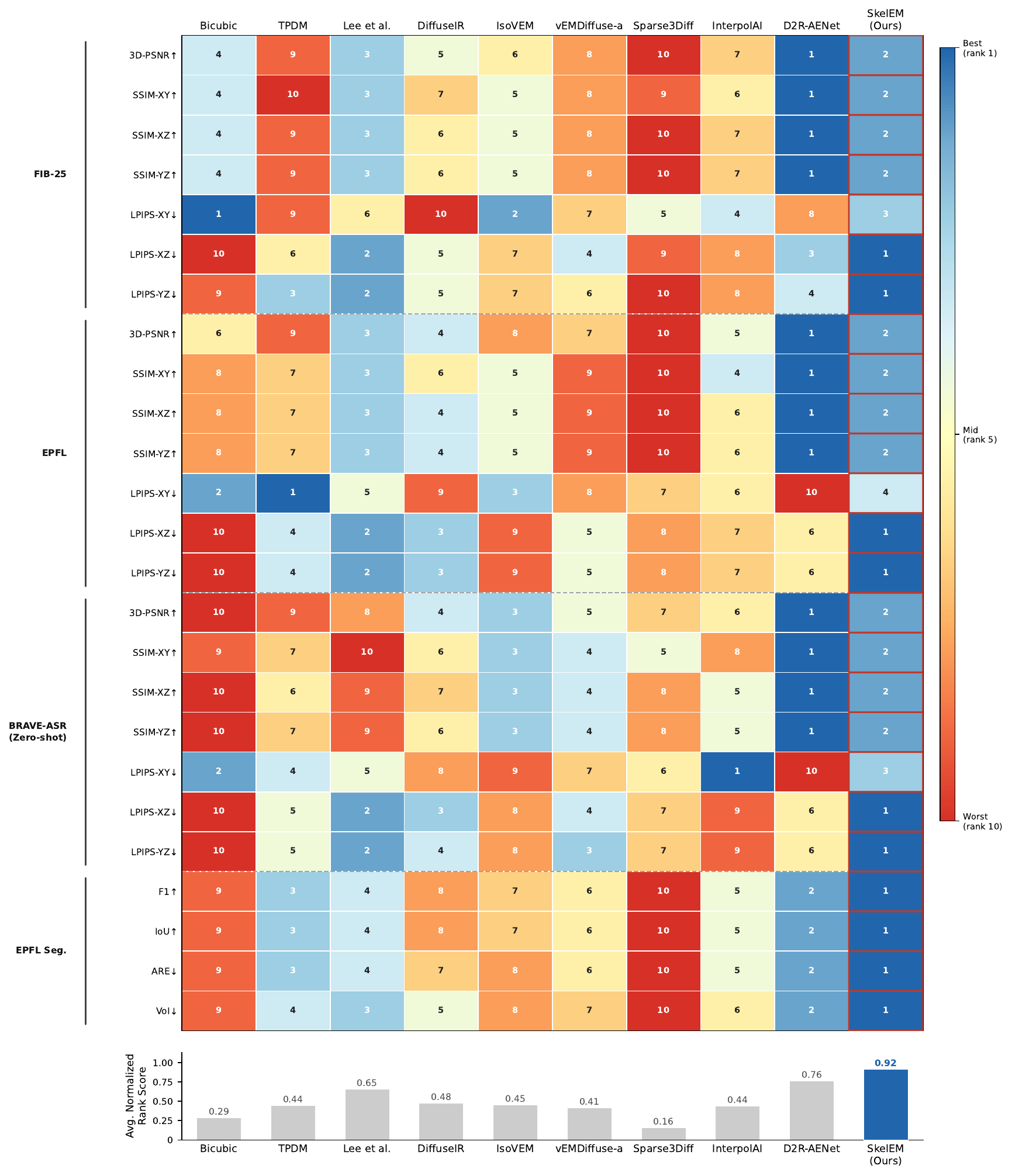}
  \caption{
  \textbf{Normalized rank heatmap across all evaluated metrics.}
  Each cell shows the ordinal rank of a method on the corresponding metric
  (1 = best, 10 = worst), mapped to a normalized score
  $s_i = 1 - (r_i - 1)/(N-1) \in [0,1]$ (blue = best, red = worst).
  For lower-is-better metrics (\textit{e.g.}~LPIPS, ARE, VoI),
  ranks are assigned in ascending order of the raw value.
  The bottom bar shows the average normalized rank score $\bar{s}$
  over all $M{=}25$ metrics, summarizing each method's overall balance
  across fidelity, perceptual quality, zero-shot generalization,
  and downstream segmentation.
  SkelEM achieves the highest $\bar{s}{=}0.92$, with no catastrophic
  failures on any individual metric, in contrast to regression-based
  methods (\textit{e.g.}~D2R-AENet) that rank high on PSNR/SSIM but
  poorly on LPIPS, and diffusion-based methods with the inverse pattern.
  }
  \label{fig:rank_heatmap}
\end{figure}

Complementing the summarized rank scores shown in Fig.~2 of the main paper, 
here we provide the full normalized rank heatmap covering all 25 metrics 
across FIB-25, EPFL, BRAVE-ASR (zero-shot), and EPFL membrane segmentation.
For each metric, the ordinal rank $r_i$ of every method is linearly mapped to a
normalized score $s_i = 1 - (r_i - 1)/(N-1) \in [0,1]$, so that the best-ranked
method always receives $1.0$ and the worst-ranked always receives $0.0$,
regardless of the absolute scale of the metric.
This rank-based normalization is deliberately chosen over value-based
normalization to avoid scale distortions caused by metrics with vastly different
dynamic ranges (\textit{e.g.}~3D-PSNR in dB \vs LPIPS in $[0,1]$).

The heatmap exposes the fidelity–perception trilemma in stark visual terms:
D2R-AENet occupies deep blue cells in PSNR and SSIM columns but turns
conspicuously red in the LPIPS columns, confirming its over-smoothing
tendency quantitatively.
Conversely, diffusion-based methods achieve competitive perceptual rankings but
display irregular fidelity patterns.
SkelEM (highlighted in red borders) maintains consistently blue-to-neutral
cells across all metric groups, with no catastrophic failures on any individual
metric, which is a pattern that is unique among all evaluated methods.
This balance is summarized in the bottom bar chart via the average normalized rank
score $\bar{s}$, on which SkelEM leads with $\bar{s} = 0.92$, followed by
D2R-AENet ($0.76$) and Lee \etal\ ($0.65$).
The gap between SkelEM and the second-ranked method ($\Delta\bar{s} = 0.16$) is
notably larger than the spread among the remaining eight methods ($0.16$--$0.65$
range), underscoring that the proposed training-signal decoupling is the key factor in resolving the trilemma rather than incremental improvements along a single axis.

\subsection{Additional Comparison with an Early MICCAI Baseline}
\label{sec:miccai_comparison}

For completeness, we additionally compare SkelEM against Deng~\etal~\cite{deng2020isotropic}, which extended lateral-to-axial restoration with unsupervised degradation learning. Quantitative comparison on the EPFL dataset is provided in \cref{tab:miccai_compare}. We note that the earlier IsoNet~\cite{weigert2017isotropic} paradigm of 2D CNN-based lateral-to-axial restoration is already well represented by more recent self-supervised baselines in our main comparison (\eg, DiffuseIR~\cite{pan2023diffuseir}), and is therefore not duplicated here.

\begin{table}[!tbp]
\centering
\footnotesize
\caption{\textbf{Comparison with an early MICCAI baseline on the EPFL dataset ($r{=}8$).} Best self-supervised results are in \textbf{bold}.}
\setlength{\tabcolsep}{3pt}
\resizebox{\linewidth}{!}{%
\begin{tabular}{l|c|ccc|ccc|cc}
\toprule
\multirow{2}{*}{Method}
  & \multirow{2}{*}{3D-PSNR$\uparrow$}
  & \multicolumn{3}{c|}{SSIM$\uparrow$}
  & \multicolumn{3}{c|}{LPIPS$\downarrow$}
  & \multicolumn{2}{c}{Membrane Seg.} \\
\cmidrule{3-10}
& & XY & XZ & YZ & XY & XZ & YZ & ARE$\downarrow$ & VoI$\downarrow$ \\
\midrule
Bicubic                                & 23.09 & 0.494 & 0.519 & 0.510 & \textbf{0.301} & 0.679 & 0.667 & 0.438 & 0.784 \\
Deng~\etal~\cite{deng2020isotropic}    & 23.17 & 0.496 & 0.538 & 0.527 & 0.326 & 0.488 & 0.480 & 0.424 & 0.677 \\
DiffuseIR~\cite{pan2023diffuseir}      & 24.48 & 0.506 & 0.563 & 0.556 & 0.398 & 0.365 & 0.379 & 0.314 & 0.608 \\
\textbf{SkelEM (ours)}                 & \textbf{25.56} & \textbf{0.619} & \textbf{0.653} & \textbf{0.646} & 0.305 & \textbf{0.321} & \textbf{0.330} & \textbf{0.213} & \textbf{0.481} \\
\bottomrule
\end{tabular}%
}
\label{tab:miccai_compare}
\end{table}

As shown in the table, Deng~\etal~\cite{deng2020isotropic} achieves only marginal improvements over bicubic interpolation across fidelity and perceptual metrics, with downstream segmentation performance well below modern diffusion-based baselines. SkelEM substantially outperforms across all axes, particularly on perceptual quality and downstream segmentation.

\subsection{Quantitative Validation on Volume Light Microscopy}
\label{sec:vlm_quantitative}

While the zebrafish retina experiment in Sec.~4.3 of the main paper demonstrates qualitative generalization to VLM, we provide additional quantitative validation using the publicly available CSBDeep mouse-liver volume~\cite{weigert2018content}, which was acquired with a high-NA objective and isotropic voxel sampling specifically for benchmarking. This enables a controlled $r=8$ axial degradation evaluation, with results in \cref{tab:vlm_quant}. The baselines in this comparison are methods that rank high across different metrics (\cref{fig:rank_heatmap}) \emph{and} remain computationally tractable at volumetric scale, so that cross-modality generalization is assessed against the strongest practically deployable competing approaches rather than an incidental subset.

\begin{table}[!tbp]
\centering
\footnotesize
\caption{\textbf{Quantitative VLM comparison under $r{=}8$ axial degradation on the CSBDeep mouse-liver volume.} Best/second-best self-supervised result in \textbf{bold}/\underline{underlined}.}
\setlength{\tabcolsep}{4pt}
\begin{tabular}{l|c|ccc|ccc}
\toprule
\multirow{2}{*}{Method}
  & \multirow{2}{*}{3D-PSNR$\uparrow$}
  & \multicolumn{3}{c|}{SSIM$\uparrow$}
  & \multicolumn{3}{c}{LPIPS$\downarrow$} \\
\cmidrule{3-8}
& & XY & XZ & YZ & XY & XZ & YZ \\
\midrule
SRUNet~\cite{heinrich2017deep} (Sup.) & 23.07 & 0.602 & 0.594 & 0.601 & 0.383 & 0.429 & 0.431 \\
\midrule
Bicubic                                & 21.30 & 0.524 & 0.482 & 0.479 & \underline{0.249} & 0.651 & 0.647 \\
InterpolAI~\cite{joshi2025interpolai}  & 20.79 & 0.483 & 0.428 & 0.424 & 0.293 & 0.632 & 0.633 \\
Lee~\etal~\cite{lee2024reference}      & 19.55 & 0.412 & 0.380 & 0.377 & 0.334 & \underline{0.353} & \underline{0.354} \\
vEMDiffuse-a~\cite{lu2024diffusion}    & 21.07 & 0.505 & 0.483 & 0.479 & 0.419 & 0.512 & 0.474 \\
D2R-AENet~\cite{chen2025self}          & \textbf{22.96} & \textbf{0.584} & \textbf{0.573} & \textbf{0.587} & 0.380 & 0.489 & 0.490 \\
\textbf{SkelEM (ours)}                 & \underline{22.20} & \underline{0.541} & \underline{0.557} & \underline{0.548} & \textbf{0.255} & \textbf{0.304} & \textbf{0.303} \\
\bottomrule
\end{tabular}
\label{tab:vlm_quant}
\end{table}

SkelEM achieves the best LPIPS in XZ/YZ views while maintaining competitive PSNR/SSIM, mirroring the fidelity-perception trade-off pattern observed on VEM benchmarks (Tabs.~1-2 in the main paper). This confirms that the training-signal decoupling design generalizes to fundamentally distinct imaging modalities under the same trade-off characterization.

\subsection{Computational Efficiency Analysis}
\label{sec:efficiency}

To complement the per-patch inference times reported in Tab.~5 of the main paper, we provide a comprehensive comparison of training and inference cost against representative diffusion-based ASR baselines. Training hardware is reported as in the respective original papers, while all inference measurements are conducted by us on a unified testbed—a single RTX 4090 with batch size $=1$, averaged over 2{,}500 runs at the standard $256\times256$ patch size—to ensure fair head-to-head comparison.

\begin{table}[h]
\centering
\footnotesize
\caption{\textbf{Comprehensive efficiency comparison against diffusion-based ASR baselines.} Training hardware is reported as in the respective original papers; per-patch inference time and peak GPU memory are measured by us on a unified testbed (single RTX 4090, batch size~$=1$, averaged over 2{,}500 runs at $256{\times}256$ patches). SkelEM achieves significant speedups while maintaining a comparable memory footprint.}
\setlength{\tabcolsep}{4pt}
\resizebox{\linewidth}{!}{%
\begin{tabular}{l|c|c|c}
\toprule
Method & Training Hardware\textsuperscript{$\dagger$} & Inference Time (ms)\textsuperscript{$\ddagger$}$\downarrow$ & Peak Memory (GB)\textsuperscript{$\ddagger$}$\downarrow$ \\
\midrule
TPDM~\cite{lee2023improving}              & $2 \times$ RTX 3090 & 402{,}788 & 7.89 \\
DiffuseIR~\cite{pan2023diffuseir}         & $8 \times$ V100     & 8{,}503   & 13.4 \\
Lee~\etal~\cite{lee2024reference}         & Not reported                  & 2{,}558   & 1.21 \\
vEMDiffuse-a~\cite{lu2024diffusion}       & $4 \times$ RTX 3090Ti   & 2{,}071   & 1.56 \\
\midrule
\textbf{SkelEM (ours)}                    & $2 \times$ RTX 4090 & \textbf{151}  & 1.95 \\
\bottomrule
\end{tabular}%
}\\[2pt]
{\footnotesize \textsuperscript{$\dagger$}\,As reported in the original paper. \textsuperscript{$\ddagger$}\,On our unified testbed (single RTX 4090).}
\label{tab:efficiency}
\end{table}

Under this unified testbed, SkelEM achieves substantially faster inference than all compared diffusion-based ASR methods, while maintaining a comparable memory footprint. This efficiency gain is enabled by the prior-truncated sampling strategy that reduces reverse diffusion from hundreds of steps to merely $\leq 5$, validating the practical advantage of structurally anchored truncation over standard iterative denoising. We note that absolute inference times may vary across hardware platforms, CPU performance, and volume tiling strategies; the comparison here reflects the relative efficiency advantage under matched testing conditions.

\section{Limitation Analysis}
\label{sec:limitation}
\begin{figure}[!tbp]
    \centering
    \includegraphics[width=\linewidth]{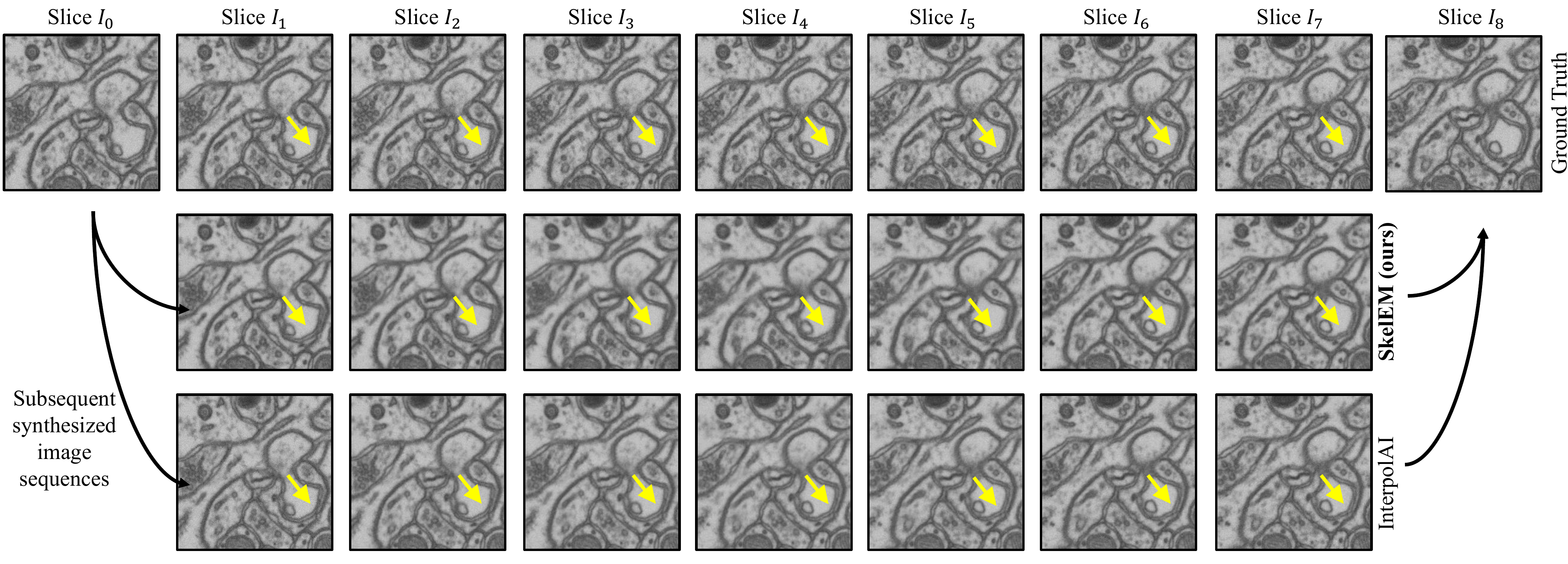}
    \caption{Limitation analysis on resolving fine double-membrane structures. Yellow arrows indicate the membrane gap. SkelEM exhibits blurring in the gap region. In contrast, InterpolAI, leveraging strong object permanence priors from natural video training, enforces a clearer geometric separation but fails to synthesize biologically realistic textures, producing scene-constant interpolations that lack structural variation.}
    \label{fig:limitation}
\end{figure}

While SkelEM excels in texture recovery, we acknowledge limitations 
in resolving fine double-membrane structures under large resampling 
factors. As visualized in \cref{fig:limitation}, SkelEM tends to 
blur the gap between nano-scale membranes, whereas InterpolAI 
maintains sharper geometric separation.

Crucially, our analysis reveals that this blurring originates from 
the skeleton generation stage. The skeleton network, which is trained 
on biological data, learns that tissue structures frequently exhibit 
complex morphological changes, such as merging or disappearing. 
Consequently, when facing the high ambiguity of a large slice gap, 
the skeleton network tends to predict a probabilistic connection. In 
contrast, InterpolAI benefits from its pre-training on large-scale 
natural videos. Natural scenes are dominated by rigid object motion 
and object permanence, leading the model to learn strong priors for 
structural continuity and boundary preservation. These priors enforce 
a hard geometric separation, even in ambiguous regions. This 
comparison highlights that natural image priors enforce strong 
geometric constraints but lack the capacity for biological texture 
synthesis.

A promising direction to resolve this tension is to initialize the 
skeleton network from a natural-video VFI model and fine-tune it on 
domain-specific biological manifolds, rather than training from 
scratch on synthetic microscopy data alone. Such an initialization 
would inherit the geometric boundary-preservation priors of natural 
video models while allowing the subsequent fine-tuning stage to adapt 
the flow estimation to the nonlinear morphological dynamics of 
biological ultrastructures. We hypothesize that this strategy could 
retain the hard geometric separation observed in InterpolAI while 
recovering the biological texture fidelity that purely natural-video 
priors cannot provide, offering a more favorable starting point for the training-signal decoupling central to SkelEM.

% \section*{Acknowledgements}
% Please insert your acknowledgments here.

% ---- Bibliography ----
%
% BibTeX users should specify bibliography style 'splncs04'.
% References will then be sorted and formatted in the correct style.
%
% \bibliographystyle{splncs04}
% \bibliography{main}
% \end{document}

\end{document}